\documentclass[lettersize,journal]{IEEEtran}
\usepackage{amsmath,amsfonts}
\usepackage{algorithmic}
\usepackage{algorithm}
\usepackage{array}
\usepackage[caption=false,font=normalsize,labelfont=sf,textfont=sf]{subfig}
\usepackage{textcomp}
\usepackage{stfloats}
\usepackage{url}
\usepackage{verbatim}
\usepackage{graphicx}
\usepackage{cite}
\usepackage{tabularx}
\usepackage{booktabs}
\usepackage{color}
\usepackage{colortbl}
\usepackage{amssymb}
\usepackage{bbm}
\usepackage{xspace}
\usepackage{multirow}
\usepackage[colorlinks,
            linkcolor=blue,       
            anchorcolor=blue,  
            citecolor=blue,   
            urlcolor=blue,
            ]{hyperref}
\usepackage{xcolor}
\usepackage{cleveref}
\usepackage{hhline}
\usepackage{vcell}
\usepackage{orcidlink} 
\begin{document}

\title{Clarity Contrast and Similarity Selection for \\ Multi-Focus Image Fusion}

\author{Yicheng Zhang$^{\orcidlink{0009-0009-0078-2472}}$, Haoyou Deng$^{\orcidlink{0009-0007-7615-7986}}$, Zhiqiang Li, Wenti Yin$^{\orcidlink{0009-0001-1875-2660}}$, Nong Sang$^{\orcidlink{0000-0002-9167-1496}}$,  \textit{Member, IEEE}, and Changxin Gao$^{\orcidlink{0000-0003-2736-3920}}$, \textit{Senior Member, IEEE}
        % <-this % stops a space
%2025.8.5注释
\vspace{10pt}
\thanks{Yicheng Zhang, Haoyou Deng, Zhiqiang Li, Wenti Yin, Nong Sang, Changxin Gao are with Key Laboratory of Image Processing and Intelligent Control, School of Artificial Intelligence and Automation, Huazhong University of Science and Technology, Wuhan, 430074, China (E-mail: $\{$yicheng$\_$zhang, haoyoudeng, zhiqiangli, yinwt, nsang, cgao$\}$@hust.edu.cn). (Corresponding author: Changxin Gao.)}
}

\markboth{Journal of \LaTeX\ Class Files,~Vol.~14, No.~8, August~2021}%
{Shell \MakeLowercase{\textit{et al.}}: A Sample Article Using IEEEtran.cls for IEEE Journals}

%2025.8.5注释


\maketitle

\begin{abstract}
Multi-focus image fusion (MFIF) aims to generate an all-in-focus image from multiple images of the same scene focused at different regions. 
%-------------------------------------------------------------------------
Most existing deep learning-based methods lack explicit interaction between the source images, which limits their performance and interpretability. 
This paper presents a novel Clarity Contrast and Similarity Selection Network (CSNet), to bridge direct information exchange for MFIF.
%-------------------------------------------------------------------------
Specifically, by contrasting the clarity differences between source images within our proposed Clarity Contrast Attention Module (CCAM), we mutually enhance sharp features while suppressing blurry ones.
This allows us to identify the exactly focused regions in each source and locate the focused-defocused boundaries.
%-------------------------------------------------------------------------
Moreover, the Defocus Spread Effect (DSE) degrades pixels in all source images around the boundaries. 
To further refine these ambiguous areas, we introduce a Similarity Selection Strategy, which reconstructs an initial clear image from source images and selects optimal pixels by comparing the similarity among them.
%-------------------------------------------------------------------------
Through this interactive approach, CSNet effectively preserves focused regions as well as recovering natural boundaries to fuse an all-in-focus output.
Extensive experiments demonstrate that our method achieves state-of-the-art performance both quantitatively and qualitatively. Our code is available on Github: https://github.com/ZYC-HUST/CSNet.
\end{abstract}


\section{Introduction}
\label{sec:Introduction}
\IEEEPARstart{A}{} fundamental goal in photography is to capture an all-in-focus (AIF) image. However, this is often hindered by the finite depth of field (DoF) of camera lenses, since a single shot cannot concurrently focus on all the objects at varying depths. 
Objects within the focal plane are projected sharply onto the image sensor, while those outside this range appear blurred, leading to defocus degradation and reduced visual clarity in the captured result.
To tackle this inherent physical limitation, multi-focus image fusion (MFIF) provides an effective computational solution, aiming to generate an AIF output through the fusion of multiple images with different focused areas. This technique has found significant applications in various fields such as consumer electronics and microscopy imaging.

%-------------------------------------------------------------------------
\begin{figure}[t]  
  \centering  
  \includegraphics[width=0.488\textwidth]{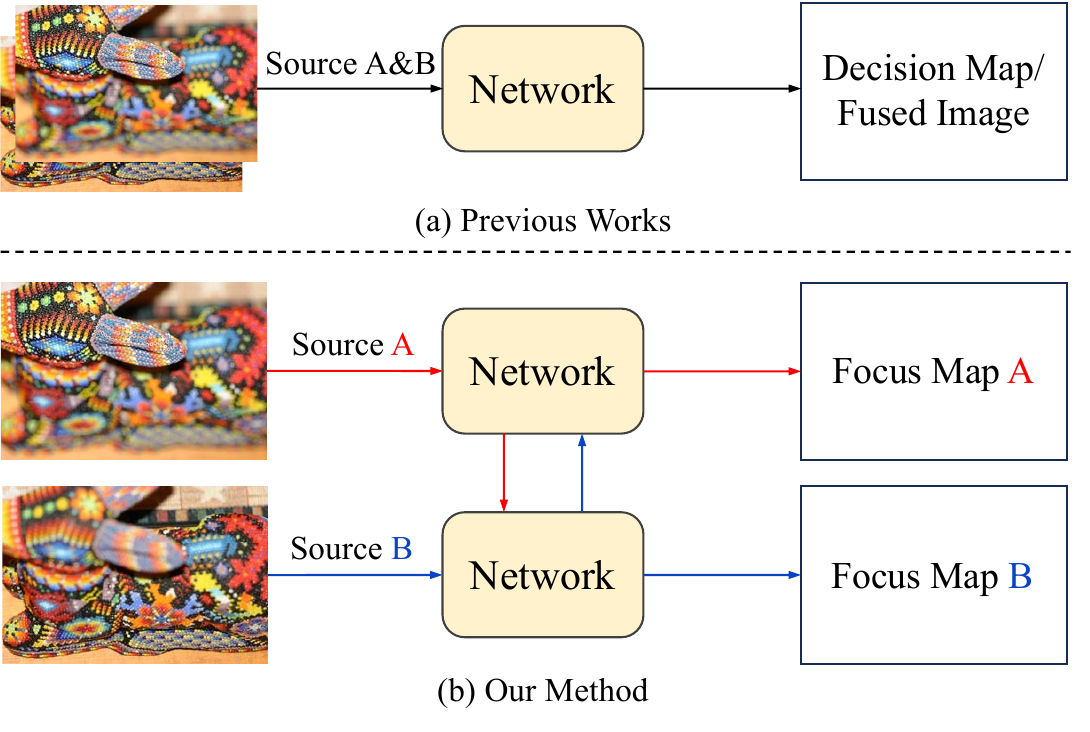}  
  \vspace{-5pt}
  \caption{Illustration of our motivation. (a) Most of the previous works process the source images together through the network. (b) Our method establishes an explicit interaction between the source images to detect focused regions and locate boundaries.} 
  \vspace{-5pt}
  \label{fig:1}  
\end{figure}
%-------------------------------------------------------------------------
Multi-focus image fusion originated in the 1980s and diverse algorithms have been developed. Conventional MFIF approaches can generally be divided into spatial domain-based~\cite{huang2018algebraic, bouzos2019conditional} and transform domain-based methods~\cite{li2013imagefusion, liu2016image}.
Nevertheless, these methods typically rely on hand-crafted features and fusion rules, which constrain their ability to generalize complex real-world scenarios.

In recent years, deep learning has been widely applied to MFIF, commonly falling into two categories: decision-based and reconstruction-based approaches. 
Decision-based methods learn a decision map by classifying pixels in the source images as either in-focus or out-of-focus, and then select the appropriate pixel to assemble the final composite~\cite{zhao2021depth, chen2021multi, nie2022mlnet}.
Reconstruction-based methods employ end-to-end networks to extract features directly from the source images and reconstruct the AIF output~\cite{li2018densefuse, quan2024dual, li2024model}.
However, the former methods normally struggle with the focused-defocused boundaries (FDB) whose focus attributes are ambiguous to distinguish, while the latter ones often suffer from detail loss in other regions away from the boundaries.
Furthermore, most of the previous works generally concatenate the source images and process them together through a holistic network, without modeling explicit interactions between source images.
These black-box working mechanisms easily lead to information losses and compromises their interpretability.

To address the above issues, we explore solutions grounded in the physical causes of defocus blur and relationship between the source images. The inputs to MFIF are a set of source images captured with different focal settings, we consider a typical scenario that involving two source images $(\boldsymbol{I}_A, \boldsymbol{I}_B)$. 
Due to the limited DoF of camera, $\boldsymbol{I}_A$ and $\boldsymbol{I}_B$ can sharply render only parts of the field, while regions outside the focal range cannot converge to clear points and are inevitably blurred. 
With various settings, their focused regions usually correspond to different depth layers. 
This imaging property suggests that in most non-boundary areas, corresponding pixels in source images tend to show complementary focus characteristics. 
Therefore, it will be beneficial to distinguish focused regions by interactively exploiting the judgment result from each other.
Such an inverse relationship can be interpreted as a negative correlation in the clarity between source images, which can be effectively characterized by a contrastive analysis.
Specifically, for a given pixel, if the clarity of $\boldsymbol{I}_A$ is significantly stronger than that of $\boldsymbol{I}_B$, it is reasonable to infer that $\boldsymbol{I}_A$ is in focus at this location.
Conversely, when $\boldsymbol{I}_A$ exhibits an ‌inferior clarity property, $\boldsymbol{I}_B$ is more likely to contain the focused information here.
The same conclusion also holds symmetrically when the relative clarity relationship is reversed. 
Through the contrast of clarity, the distinctly focused pixels can be identified and are suitable for direct preservation to be output.

However, in cases where the clarity levels of $\boldsymbol{I}_A$ and $\boldsymbol{I}_B$ are similar, both may be either focused or defocused, which frequently arise near the boundaries. Since they are difficult to distinguish and both images may be degraded here by the Defocus Spread Effect (DSE), directly extracting pixels from the source images is no longer a reliable solution, instead easily introduce unnatural transitions or artifacts, and further decision and improvement by additional means are necessary.
Considering the superiority of reconstruction-based methods in handling such ambiguous regions, reconstructing these areas provides a more effective solution to generate a pleasing transition across different fusion parts.

Motivated by the preceding insight, we propose a novel Clarity Contrast and Similarity Selection Network (CSNet) to establish an explicit and effective interaction for MFIF.
%-------------------------------------------------------------------------
Specifically, a Clarity Contrast Attention Module (CCAM) is designed to reciprocally identify the precisely focused regions in the source images $\boldsymbol{I}_A$ and $\boldsymbol{I}_B$.
Within the CCAM, the discrepancy in clarity between $\boldsymbol{I}_A$ and $\boldsymbol{I}_B$ is utilized as weights to inversely modulate each other, emphasizing sharp features while inhibiting blurry ones.
After the interactions on multi-scale, we generate two focus maps that pinpoint the pixels from source images to be preserved directly for output.
%-------------------------------------------------------------------------
Simultaneously, for the remaining areas where the clarity of $\boldsymbol{I}_A$ and $\boldsymbol{I}_B$ are both attenuated due to the Defocus Spread Effect, we classify them as the focused-defocused boundaries.
%-------------------------------------------------------------------------
To further refine the uncertain parts lying outside the focusable range of sources, a Similarity Selection Strategy is introduced.
In particular, we reconstruct an initial clear image $\boldsymbol{I}_C$ from the source images and measure perceptual distance among them within the feature space of a VGG-16 network to compare the similarity.
Crucially, we select pixels at boundaries from source images only if the corresponding positions in $\boldsymbol{I}_C$ show substantial bias toward one of $\boldsymbol{I}_A$ and $\boldsymbol{I}_B$, which helps to enhance the fidelity of fusion.
Otherwise, the reconstructed results from $\boldsymbol{I}_C$ are utilized for output to alleviate the DSE.
%-------------------------------------------------------------------------
Finally, according to the focus maps and selection strategy, proper pixels are fused to achieve the AIF output. 

In summary, our paper makes the following contributions:
\begin{itemize}
    \item We propose the CSNet to build an explicit interaction for MFIF. This approach enables the collection of focused regions from multiple source images jointly with restoration of boundaries to fuse an AIF output.
    \item We present the Clarity Contrast Attention Module, which interactively detects the focused regions and accurately locates the boundaries by contrasting the clarity differences between the source images. 
    \item We introduce the Similarity Selection Strategy to polish the boundary areas. It generates an initial clear image from the source images and adopts source or reconstructed pixels by comparing the similarity among them.
    \item Extensive experiments on four benchmarks validate the superiority of our CSNet in both evaluation metrics and visual quality compared to state-of-the-art approaches.
\end{itemize}

%-------------------------------------------------------------------------
\begin{figure}[t]  
  \centering  
  \includegraphics[width=0.488\textwidth]{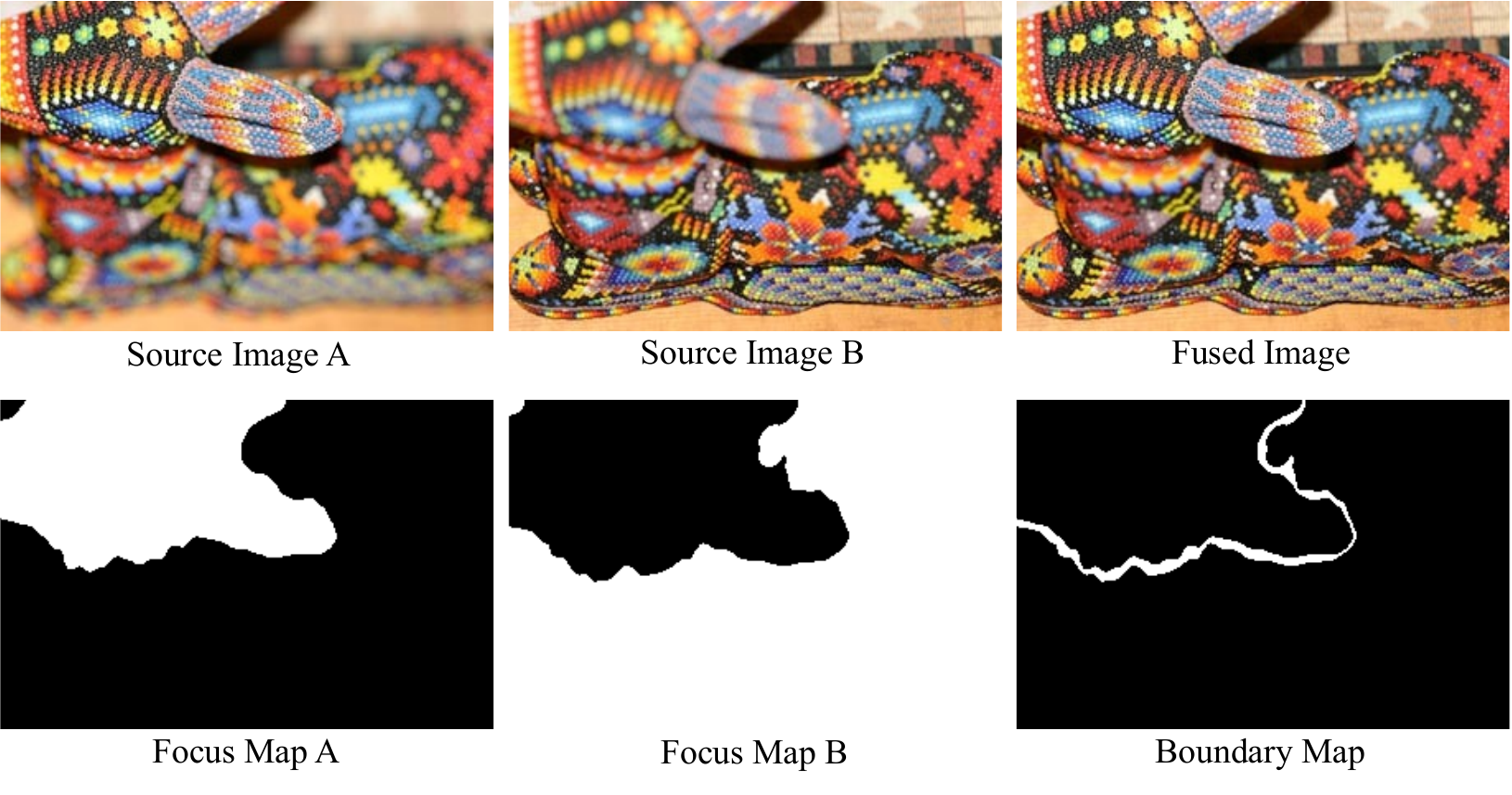}  
  \vspace{-10pt}
  \caption{An example of MFIF task and our method.
  The first row shows source images focusing on the foreground and background, respectively, along with the fused result from the proposed method. 
  The second row displays binary masks for the focused regions and the boundaries in the source images, which are generated by our CSNet.} 
  \vspace{-10pt}
  \label{fig:2}  
\end{figure}
%-------------------------------------------------------------------------
%-------------------------------------------------------------------------
\begin{figure*}[t]  
  \centering  
  \includegraphics[width=1.0\textwidth]{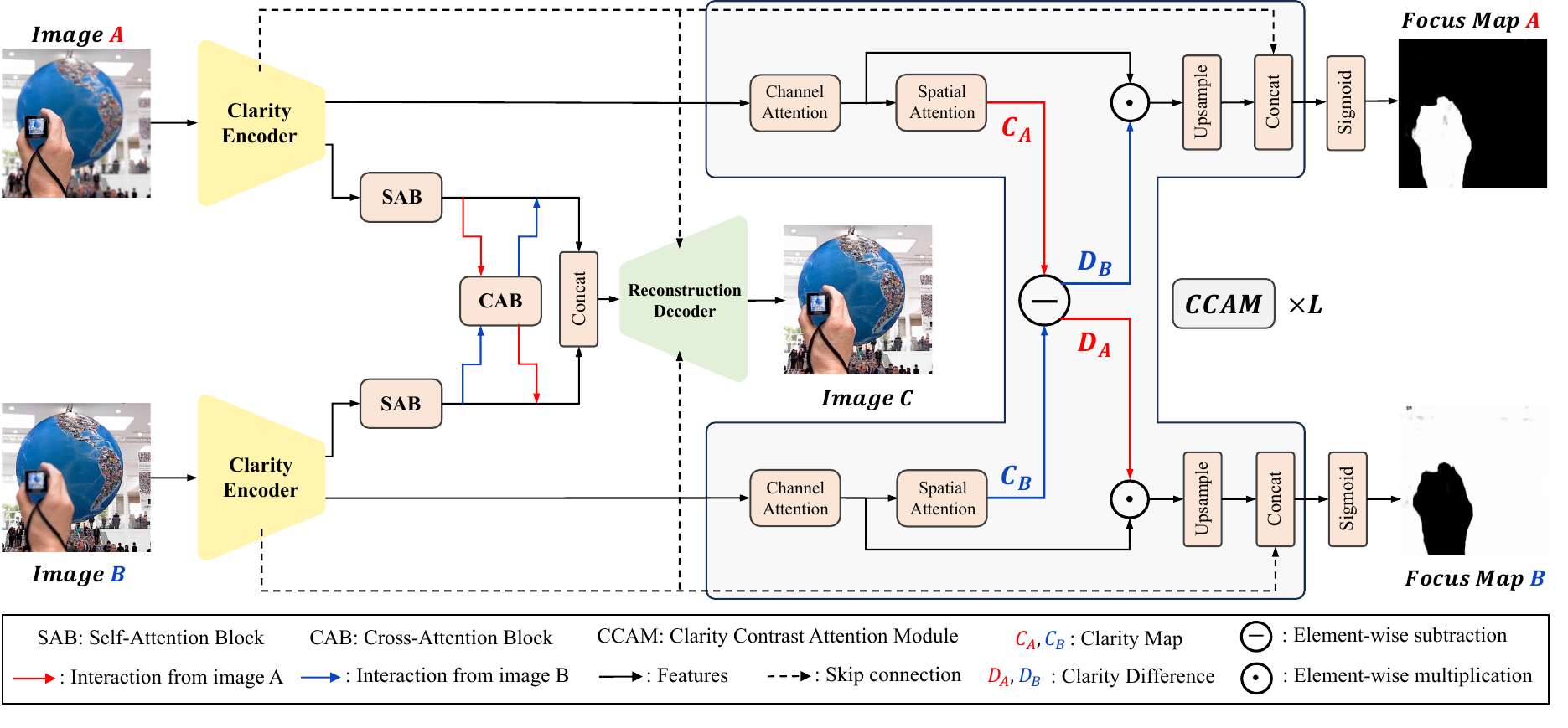}  
  \vspace{-10pt}
  \caption{The architecture of our CSNet. Given the source image pair, siamese-style encoders are first utilized to extract clarity features. 
  Subsequently, we employ CCAMs to contrast the clarity of source images and inversely modulate each other on multi-scale, generating two focus maps to identify the precisely focused regions. 
  Concurrently, the extracted features are enhanced through cross-attention to reconstruct an initial clear image for later boundary refinement.} 
  \vspace{-5pt}
  \label{fig:3}  
\end{figure*}
%-------------------------------------------------------------------------

\section{Related Work}
\label{sec:Related Work}
\subsection{Traditional MFIF}
Conventional MFIF algorithms are broadly categorized into spatial domain-based and transform domain-based methods~\cite{liu2020multi}. 
Spatial domain-based methods operate directly on the source images and can be further divided into three types pixel-based~\cite{li2013image}, block-based~\cite{guo2015high}, and region-based~\cite{li2006region} approaches according to the size of the processing units. 
In these approaches, a focus map is constructed to indicate the focus status of each pixel by evaluating the activity levels of the source images.
By contrast, transform domain-based methods operate by mapping source images into a transform domain, integrating the resulting coefficients with predefined fusion strategies, and reconstructing the fused image through an inverse transform. 
They including sparse representation methods~\cite{zhang2018robust}, multi-scale methods~\cite{kou2018multi} and gradient domain-based methods~\cite{zhou2014multi}.
However, human-designed features cannot sufficiently capture the inherent properties of images, and the limitations of hard-rules further hinder fusion performance.

%-------------------------------------------------------------------------

\subsection{Deep-Learning Based MFIF}
In recent years, deep learning has emerged as a dominant paradigm, with existing methods broadly grouped into decision-based and reconstruction-based methods~\cite{zhang2021deep} depending on their fusion strategies. Moreover, an increasing number of studies have investigated joint frameworks. 
\vspace{5pt}
%-------------------------------------------------------------------------

\noindent \textbf{Decision-Based Methods} 
These methods usually predict a focus map to determine the focus property of each pixel, and then obtain a pixel-level binary classification decision map to select the corresponding source image pixels. 
Liu et al.~\cite{liu2017multi} pioneered the use of CNNs to generate decision maps;
MLFCNN~\cite{yang2019multilevel} is a multi-level features convolutional neural network; 
GEU-Net~\cite{xiao2020global} designs a feature pyramid to improve discriminative power;  
MADCNN~\cite{lai2019multi} employs a visual attention unit to help the network locate the focused region more accurately; 
EAY-Net\cite{wang2023multi} introduce edge-preservation techniques to alleviate the influence of inaccurate boundaries with burrs in decision maps.
\vspace{5pt}

%-------------------------------------------------------------------------
\noindent \textbf{Reconstruction-Based Methods} 
These methods establish an end-to-end mapping from the source images to the fused output, often following a three-stage ``extraction-fusion-reconstruction" pipeline. 
IFCNN~\cite{zhang2020ifcnn} firstly introduces an end-to-end network; 
Xu et al.~\cite{xu2018multi} proposed the first two-stream MFIF method, which processes one source image using individual branch; 
SwinFusion~\cite{ma2022swinfusion} combines CNN with transformer to efficiently capture both local and global information; 
SFIMFN~\cite{huang2024efficient} integrates high-low frequency information from the spatial and frequency domains; 
Fusion2Void~\cite{lin2024fusion2void} ingeniously tackles the challenge of missing ground-truth by framing image inpainting as an auxiliary task. 
\vspace{5pt}
%-------------------------------------------------------------------------

\noindent \textbf{Joint Methods} 
Both types of deep learning-based methods have their strengths and weaknesses. Although the decision-based methods do well in preserving the in-focus details, they often introduce artifacts in focused-defocused boundaries (FDB). In contrast, reconstruction-based methods generate smoother fusion boundaries, but typically suffer from detail loss in other non-boundary regions.
To harness the complementary advantages of them, several joint approaches have been proposed. Liu et al.~\cite{liu2022multi} developed a two-stage DNN, which first recovers an initial fused image and then refines it via a decision module; GRFusion~\cite{li2023generation} integrating the advantage of feature reconstruction and pixel recombination into a single fusion framework by detecting hard cases; In the DB-MFIF~\cite{zhang2024exploit}, an end-to-end branch and a decision based branch are proposed to mutually assist each other; DMANet~\cite{quan2025multi} estimates defocus blur based on the physical model to combine deblurring with decision-making.
\vspace{-5pt}
%-------------------------------------------------------------------------

\subsection{Interactive Approaches}
Recent low-level vision approaches have explored feature interaction to improve the quality of images. 
CAT~\cite{chen2022cross} aggregates image-restoration features across spatial windows, and HAT~\cite{chen2023activating} introduces hybrid attention with overlapping cross-attention for image super-resolution.
In image fusion, Text-IF~\cite{yi2024text} leverages semantic interaction for degradation-aware process, while CDDFuse~\cite{zhao2023cddfuse} introduces correlation-driven feature decomposition to separate modality-shared and modality-specific information.
DADA~\cite{yan2020disparity} exploits stereo correspondence to inject disparity priors for stereo image restoration, Steformer~\cite{lin2023steformer} model cross-view dependencies through parallax attention.
For MFIF, most of existing methods are generally constrained to unified processing of the source images. 
Although some methods have considered focus relationships: 
DRPL~\cite{li2020drpl} creates complementary focus maps using a pairwise strategy, 
DTMNet~\cite{xiao2021dtmnet} employs discrete Tchebichef moments for focus discrimination,
FRP~\cite{liu2023focus} designs relationship perception for unsupervised methods; 
explicitly formulation for inter-source interaction still remains unexplored.

\section{Method}

%-------------------------------------------------------------------------
\begin{figure*}[t]  
  \centering  
  \includegraphics[width=1\textwidth]{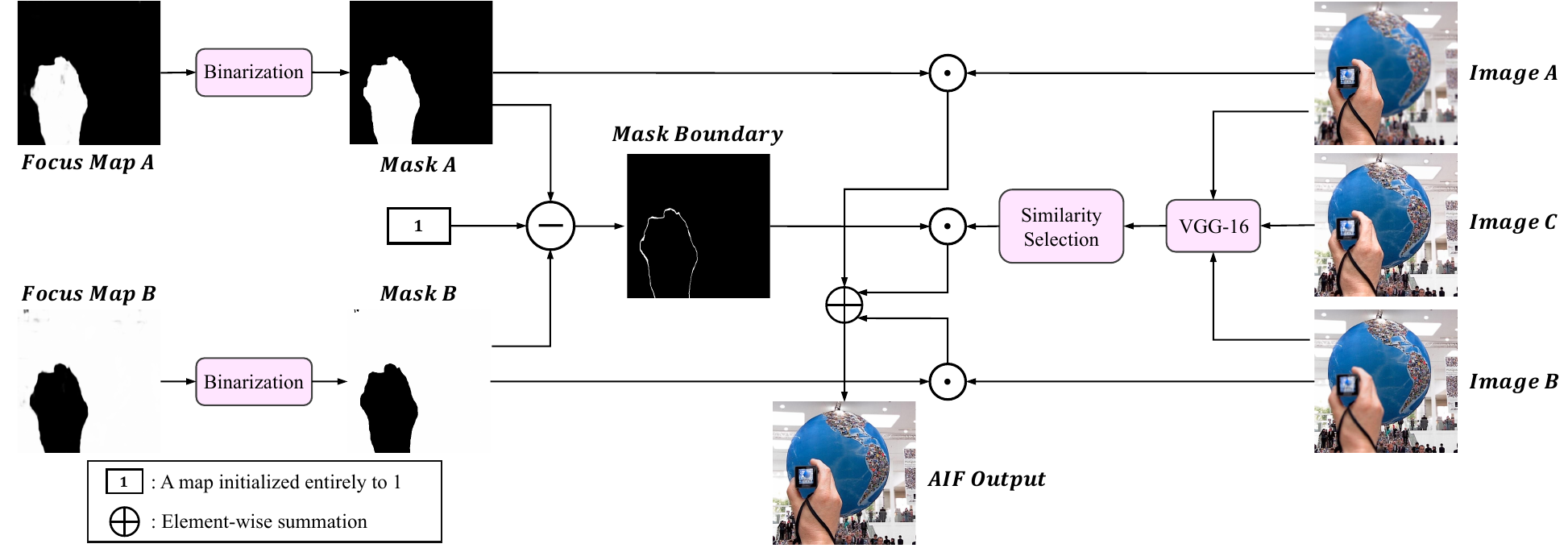}  
  \vspace{-5pt}
  \caption{Illustration of our Similarity Selection Strategy. The focus maps identify the focused regions in the source images, which should be directly preserved. 
  For the boundaries, we measure similarity between source images and reconstructed image by computing the Euclidean distance between the feature maps. 
  Then pixels are selected from one of the source image whose content exhibits significantly higher similarity to the initial clear image at the corresponding location; otherwise, pixels from the reconstructed result are chosen.} 
  \vspace{-5pt}
  \label{fig:4}  
\end{figure*}
%-------------------------------------------------------------------------

\subsection{Overview}
An overview of the proposed CSNet framework is shown in Figure~\ref{fig:3}. 
Without loss of generality, we consider the basic case of MFIF with two input images $(\boldsymbol{I}_A, \boldsymbol{I}_B)\in\mathbb{R}^{H \times W \times 3}$, which are captured with different focal planes. 
The framework can be naturally extended to handle more input images by processing them in a staged pairwise manner.
                                                              
We begin by leveraging clarity encoders to extract features from every source image, capturing both local texture details and global contextual information.
To ensure the consistency of extraction, they share the same weights. 
Specifically, the encoder consists of $L$ hierarchical layers, $L = 4$ in our network, each including a convolutional block and a residual block. 
The number of channels for layers are [16, 32, 64, 128], and after each layer, the resolution is halved.

Following that, the clarity features pass through the Clarity Contrast Attention Module (CCAM) at every level to achieve a coarse-to-fine process~\cite{li2023model}. By contrasting the clarity differences between source images, CCAM mutually highlights clear representations while diminishing blurry ones.
After $L$ rounds of such interactions, the confidently focused areas in $\boldsymbol{I}_A$ and $\boldsymbol{I}_B$ are delineated by two focus maps $\boldsymbol{M}_A$ and $\boldsymbol{M}_B\in[0, 1]^{H \times W}$, which indicate the likelihood of each pixel being in focus.
In parallel, to improve the boundaries whose focus properties are ambiguous in both source images, the extracted features undergo self-attention to enhance contextual representations, followed by cross-attention to facilitate information exchange between the inputs. 
This process yields an initial reconstructed image $\boldsymbol{I}_C \in \mathbb{R}^{H \times W \times 3}$, which has been demonstrated in prior work~\cite{zhang2021deep, zhang2024exploit} to produce natural and seamless transitions at the focused-defocused boundaries.

Finally, a novel fusion strategy is applied to achieve the AIF output $\boldsymbol{I}_{F}\in\mathbb{R}^{H \times W \times 3}$, as shown in Figure~\ref{fig:4}. 
We directly reserve the exactly focused regions in the source images for output, while selecting pixels from proper image at the FDB areas through similarity comparison by measuring perceptual distance within the feature space of a pre-trained VGG-16 network~\cite{zhang2018unreasonable}.

\subsection{Clarity Contrast Attention Module}
To build an explicit interaction between the source images and accurately localize focused regions, we design a Clarity Contrast Attention Module (CCAM) to contrast their clarity across multiple feature levels. To align with the hierarchical structure of the encoder, $L$ CCAMs are embedded into the network, enabling progressive clarity-aware interaction from coarse to fine scales.

Within CCAM, to emphasize features closely related to clarity, such as edges, chrominance and luminance, we first apply channel attention to adaptively filter and amplify salient channels containing clarity-relevant information.  
Subsequently, a spatial attention mechanism is employed to estimate the positional significance of features, where regions with stronger magnitude of clarity are assigned higher attention, thereby generating two clarity maps to characterize the spatial distribution.
Finally, the discrepancy between the two clarity maps is computed and serves as a set of weighting factors to reciprocally modulate the channel-attended features. 
The detailed design of this module is described as follows.

\begin{figure}[t]  
  \centering  
  \includegraphics[width=0.485\textwidth]{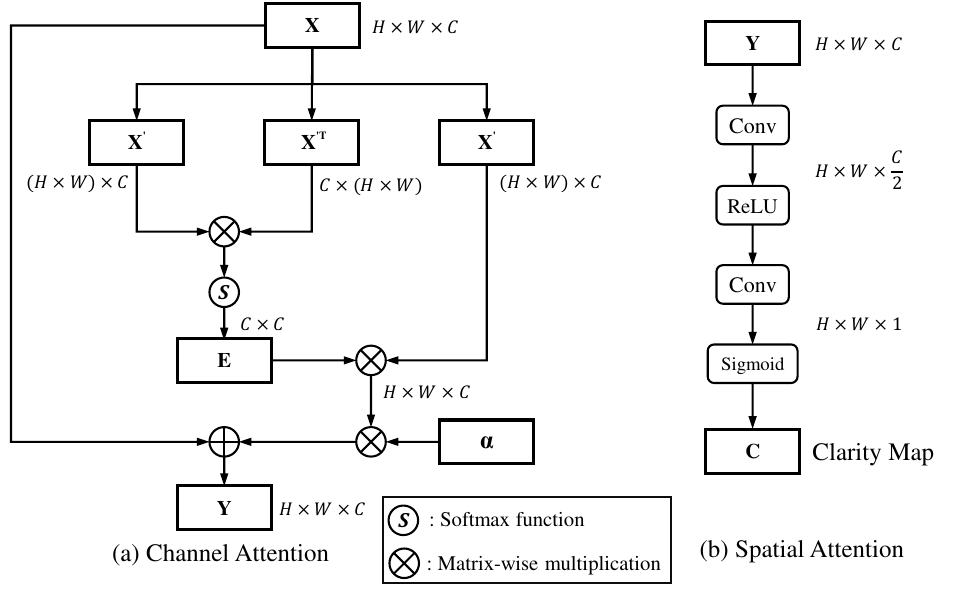}  
  \vspace{-10pt}
  \caption{Details of the channel attention module and spatial attention module.}
  \vspace{-5pt}
  \label{fig:5}  
\end{figure}

The channel attention module is illustrated in Figure~\ref{fig:5} (a). Given an arbitrary feature $\boldsymbol{X}\in\mathbb{R}^{H \times W \times C}$, we first reshape it to $\boldsymbol{X}^{'}\in\mathbb{R}^{N \times C}$, and perform a matrix multiplication between $\boldsymbol{X}^{'}$ and its transpose $\boldsymbol{X}^{'T}$. Afterwards, the attention map $\boldsymbol{E}\in\mathbb{R}^{C \times C}$ is generated by applying a Softmax function:
\begin{equation}
e_{ij} = \frac{ \exp( x_i \cdot x_j ) }{ \sum_{i=1}^{C} ( \exp( x_i \cdot x_j ) ) },
\label{eq:1} 
\end{equation}
where $e_{ij}$ measures the influence of the i-th channel on the j-th channel. 
A larger value of $e_{ij}$ indicates a stronger semantic or structural correlation between the two channels, implying that information from the $i$-th channel is more beneficial for enhancing the representation of the $j$-th channel.

Based on the learned channel attention map, each channel is reconstructed as a weighted aggregation of all input channels. 
To prevent the attention operation from disrupting the original feature distribution at the early training stage, we introduce a learnable scale factor $\alpha$ and combine the attention-enhanced features with the original input through a residual connection:
\begin{equation}
\boldsymbol{Y}_{j} = \alpha \sum_{i=1}^{C} (e_{ji} \boldsymbol{X}_{i}) + \boldsymbol{X}_{j},
\label{eq:2} 
\end{equation}
where $\boldsymbol{Y}\in\mathbb{R}^{H \times W \times C}$ denotes the output feature. 
The scale factor $\alpha$ is initialized to zero, which makes the module initially behave as an identity mapping and preserves the stability of the pre-attention representation. 
As training proceeds, $\alpha$ is gradually optimized to control the contribution of the channel-wise contextual aggregation. 

It can be inferred from Eq.\ref{eq:2} that the the resulting feature $\boldsymbol{Y}$ is formulated as a residual aggregation of the original feature and a weighted combination of all channels.This enables the module to capture long-range dependencies across channels and adaptively recalibrate feature responses.
In particular, channels that are strongly associated with image clarity reinforce one another, enabling adaptive enhancement of clarity-relevant responses while less informative or redundant channels are suppressed, improving clarity representation, and strengthening discriminative capability.

As shown in Figure~\ref{fig:5} (b), the spatial attention module consists of two $3 \times 3$ convolutional layers and a Sigmoid activation, producing a single-channel clarity map $\boldsymbol{C}\in[0, 1]^{H \times W}$, which locates the most informative features in the spatial domain. 
In this context, spatial locations containing sharp edges, fine textures, and prominent structural details are assigned higher responses, indicating a higher likelihood of being in focus. 
Conversely, blurred or less informative regions tend to receive lower attention. 
Thus, the clarity map $\boldsymbol{C}$ provides an spatial characterization of clarity distribution and serves as a reliable cue for subsequent clarity contrast.
%-------------------------------------------------------------------------
%-------------------------------------------------------------------------

Furthermore, we take the features $\boldsymbol{X}_{A}^{(k)}$ and $\boldsymbol{X}_{B}^{(k)}$ at the $k$-th level of CCAMs as examples to illustrate the realization of clarity contrast and inverse modulation.
After obtaining the channel-attended features $\boldsymbol{Y}_{A}^{(k)}$ and $\boldsymbol{Y}_{B}^{(k)}$, together with their corresponding clarity maps $\boldsymbol{C}_{A}^{(k)}$ and $\boldsymbol{C}_{B}^{(k)}$, we quantify the contrast in clarity between the two source images by measuring the discrepancy of their clarity maps:
\begin{equation}
\boldsymbol{D}_{A}^{(k)} = \boldsymbol{C}_{A}^{(k)} - \boldsymbol{C}_{B}^{(k)}, 
\boldsymbol{D}_{B}^{(k)} = \boldsymbol{C}_{B}^{(k)} - \boldsymbol{C}_{A}^{(k)},
\end{equation}
where $\boldsymbol{D}_{A}^{(k)}$ denotes the clarity advantage of $\boldsymbol{I}_A$ over $\boldsymbol{I}_B$, serving as weights to modulate the features originating from $\boldsymbol{I}_B$, while $\boldsymbol{D}_{B}^{(k)}$ plays a symmetric role for $\boldsymbol{I}_A$.
In this way, inverse modulation is established based on their inherent negative correlation, encouraging complementary feature enhancement between the two branches.

Specifically, for a given spatial position, a positive value in $\boldsymbol{D}_{A}^{(k)}$ indicates that $\boldsymbol{I}_A$ exhibits a higher clarity response than $\boldsymbol{I}_B$, suggesting that $\boldsymbol{I}_B$ is likely defocused and should be suppressed.
In case the value of $\boldsymbol{D}_{A}^{(k)}$ approaches zero, it means the clarity of $\boldsymbol{I}_A$ and $\boldsymbol{I}_B$ present comparable, so no significant impact is imposed on $\boldsymbol{I}_B$. 
In contrast, a negative value reveals that the possibility of $\boldsymbol{I}_B$ being in-focus at this location is reliably high, whose feature are accordingly enhanced. 
Similarly, $\boldsymbol{D}_{B}^{(k)}$ can be delivered for equivalence modulation on $\boldsymbol{I}_A$. The overall process can be formulated as:
\begin{equation}
\begin{aligned}
\boldsymbol{Y}_{A}^{(k)'} &= \boldsymbol{Y}_{A}^{(k)} \odot (\boldsymbol{1}-\boldsymbol{D}_{B}^{(k)}) \\
\boldsymbol{Y}_{B}^{(k)'} &= \boldsymbol{Y}_{B}^{(k)} \odot (\boldsymbol{1}-\boldsymbol{D}_{A}^{(k)})
\end{aligned},
\end{equation}
where $\boldsymbol{1}$ is a map initialized entirely to $1$, used to model the effects of negative correlation.

In the end, the modulated features are upsampled and concatenated with the encoded features from the next layer, forming the input for the $k+1$-th level of CCAMs:
\begin{equation}
\begin{aligned}
\boldsymbol{X}_{A}^{(k+1)} &= Concat[Up(\boldsymbol{Y}_{A}^{(k)'}),  \boldsymbol{F}_{A}^{(k+1)}]\\
\boldsymbol{X}_{B}^{(k+1)} &= Concat[Up(\boldsymbol{Y}_{B}^{(k)'}),  \boldsymbol{F}_{B}^{(k+1)}]
\end{aligned},
\end{equation}
This allows us first utilize the high-level semantics of deep features to ensure correct region classification, followed by progressively integrating the sharp details from shallow features. 
After $L$ rounds of iterations in the CCAMs, the final features pass through a Sigmoid function respectively to obtain two focus maps $\boldsymbol{M}_A$ and $\boldsymbol{M}_B\in[0, 1]^{H \times W}$.

The focus maps pinpoint the definitively focused regions in the source images; but there still remains some regions whose focus attributes are hard to classify, typically occurring near the FDB. We can get the boundary map $\boldsymbol{M}_{C}\in[0, 1]^{H \times W}$ to localize these areas via the below calculation:
\begin{equation}
\boldsymbol{M}_{C} = \boldsymbol{1} - \boldsymbol{M}_{A} - \boldsymbol{M}_{B}.
\end{equation}

%-------------------------------------------------------------------------
\subsection{Cross-Reconstruction}
Considering the advantage of the reconstruction-based methods in modeling smooth and coherent boundaries, we reconstruct an initial clear image $\boldsymbol{I}_C$ from the source images to facilitate subsequent boundary refinement. 
To further realize our idea of the interaction, we introduce a cross-attention mechanism to achieve this process~\cite{li2024crossfuse, li2025enhanced}. 
Given the clarity-aware features from the deepest encoder layer, we first employ a self-attention block (SAB)~\cite{fan2021multiscale} to capture long-range spatial dependencies, allowing each feature to emphasize its own crucial parts. 

Following that, the enhanced features $\boldsymbol{E}_{A}$ and $\boldsymbol{E}_{B}$ are then fed into a cross-attention block (CAB)~\cite{chu2022nafssr} to exchange complementary information. 
Different from standard cross-attention, which mainly performs generic feature matching between two inputs, our cross-attention block is designed as a focus-aware reciprocal interaction mechanism. 
It enables each source branch to actively retrieve complementary focus evidence from its counterpart while preserving its own intra-view structural representation. 
Specifically, in each interaction direction, the $query$ matrix is projected from the source inter-view feature, whereas the $key$ and $value$ matrices are derived from the target intra-view feature. 
%This asymmetric design allows one branch to selectively search for structurally relevant and complementary information from the other, making it particularly effective for ambiguous boundary regions where focus cues are difficult to determine from a single source.

Through this reciprocal formulation, two bidirectional interaction flows, $\boldsymbol{I}_{A\rightarrow B}$ and $\boldsymbol{I}_{B\rightarrow A}$, are obtained to encode complementary focus cues transferred across the source images. 
Rather than directly replacing the original representations, these interaction features are injected into the corresponding branches through learnable residual enhancement:
\begin{equation}
\begin{aligned}
\boldsymbol{E}_{A}^{'} &= \beta_1 \cdot \boldsymbol{I}_{B\rightarrow A}+ \boldsymbol{E}_{A}, \\
\boldsymbol{E}_{B}^{'} &= \beta_2 \cdot \boldsymbol{I}_{A\rightarrow B}+ \boldsymbol{E}_{B},
\end{aligned}
\end{equation}
where $\beta_1$ and $\beta_2$ are learnable channel-wise scaling factors. 
They adaptively regulate the strength of cross-view information injected into each channel, enabling the network to exploit complementary focus cues while avoiding excessive interference from unreliable responses. 
Consequently, the enhanced representations $\boldsymbol{E}_{A}^{'}$ and $\boldsymbol{E}_{B}^{'}$ integrate intra-view contextual consistency with inter-view focus complementarity. 

To further strengthen the reconstruction process, we introduce a focus-guided interactive decoding strategy. 
Instead of directly passing skip-connected encoder features to the decoder, the cross-enhanced representation is used to regulate the shallow features before their injection. 
This is because shallow features preserve abundant spatial details, but they may also contain defocus-contaminated responses. 
Therefore, adaptive filtering is necessary to prevent blurred structures from being propagated into the reconstructed image.

Specifically, the cross-enhanced features are first concatenated and projected into a unified reconstruction representation through a convolutional layer:
\begin{equation}
\boldsymbol{F}_{C} = Conv(Concat[\boldsymbol{E}_{A}^{'} , \boldsymbol{E}_{B}^{'}]),
\end{equation}
At the $n$-th decoding stage, a guidance map $\boldsymbol{M}_{G}^{(n)}$ is generated by processing the upsampled decoder feature:
\begin{equation}
\boldsymbol{M}_{G}^{(n)} = \sigma(Conv(Up(\boldsymbol{F}_{C}^{(n+1)})),
\end{equation}
where $\sigma(\cdot)$ is a Sigmoid function. 
The skip features are then gated by the guidance map and subsequently fused with the decoder feature for progressive reconstruction:
\begin{equation}
\boldsymbol{F}_{C}^{(n)} = Up(\boldsymbol{F}_{C}^{(n+1)}) + \boldsymbol{M}_{G}^{(n)} \odot Concat[\boldsymbol{F}_{A}^{(n)}, \boldsymbol{F}_{B}^{(n)}],
\end{equation}
where $\boldsymbol{F}_{A}^{(n)}, \boldsymbol{F}_{B}^{(n)}$ are skip features from the $n$-th encoder.
Through this design, the decoder selectively absorbs clarity-consistent details from shallow layers rather than indiscriminately reusing all encoder features. 
As a result, sharp structures can be better preserved, while defocus-related responses are suppressed during reconstruction, leading to a more coherent and reliable initial clear image $\boldsymbol{I}_C$.

%-------------------------------------------------------------------------
\begin{table*}[ht]
  \centering
  \caption{Quantitative comparison on the MFFW and Lytro datasets. The best and second results are highlighted by \textbf{bold} and \underline{underline}.}
  \normalsize
  \setlength{\heavyrulewidth}{1.2pt}
  \setlength{\lightrulewidth}{1pt}  
  \renewcommand{\arraystretch}{1.3}
  \resizebox{\textwidth}{!}{
  \begin{tabular}{l|cccccc|cccccc|c}
    \toprule
    \multirow{2}{*}{\textbf{Method}} & \multicolumn{6}{c|}{\textbf{MFFW}} & \multicolumn{6}{c|}{\textbf{Lytro}} & \multirow{2}{*}{\textbf{Para.(M)}} \\
    \cmidrule(lr){2-7} \cmidrule(lr){8-13}
     & $Q_{MI}$ & $Q_{NCIE}$ & $Q_{G}$ & $Q_{Y}$ & $Q_{C}$ & $Q_{CB}$ & $Q_{MI}$ & $Q_{NCIE}$ & $Q_{G}$ & $Q_{Y}$ & $Q_{C}$ & $Q_{CB}$ & \\
    \midrule
    CSR (2016) & 0.8943 & 0.8250 & 0.6982 & 0.8330 & 0.7759 & 0.6760 & 0.9724 & 0.8314 & 0.7253 & 0.9147 & 0.8673 & 0.7489 & ---- \\
    DSIFT (2016) & 1.1406 & 0.8400 & 0.7359 & 0.9430 & 0.8615 & 0.7355 & 1.1876 & 0.8454 & 0.7617 & 0.9881 & 0.9099 & \underline{0.8093} & ---- \\
    MFM (2017) & 1.0635 & 0.8351 & 0.7359 & 0.9483 & 0.8694 & 0.7404 & 1.1601 & 0.8436 & \textbf{0.7621} & 0.9881 & \textbf{0.9114} & \underline{0.8093} & ---- \\
    SVDDCT (2017) & 1.1602 & 0.8405 & 0.7382 & 0.9594 & 0.8695 & 0.7317 & 1.1868 & 0.8455 & 0.7593 & 0.9860 & 0.9065 & 0.8039 & ---- \\
    IFCNN (2020) & 0.8206 & 0.8218 & 0.6765 & 0.8715 & 0.8190 & 0.6423 & 0.9388 & 0.8298 & 0.7301 & 0.9518 & 0.8909 & 0.7294 & 0.084 \\
    SESF (2021) & 1.0876 & 0.8364 & \underline{0.7365} & 0.9588 & 0.8742 & 0.7418 & 1.6808 & 0.8435 & 0.7601 & 0.9879 & 0.9098 & 0.8064 & 0.074 \\
    GACN (2022) & 1.0820 & 0.8363 & 0.7359 & 0.9273 & 0.8445 & 0.7192 & 1.1723 & 0.8446 & 0.7615 & 0.9878 & \underline{0.9109} & 0.8062 & 0.069 \\
    GRFusion (2023) & 1.1426 & 0.8401 & 0.7418 & 0.9334 & 0.8570 & 0.7223 & 1.1879 & 0.8454 & \underline{0.7620} & 0.9863 & 0.9090 & 0.8071 & 9.857 \\
    MUFusion (2023) & 0.7551 & 0.8194 & 0.6063 & 0.8167 & 0.7753 & 0.6205 & 0.8088 & 0.8238 & 0.6728 & 0.8997 & 0.8453 & 0.6819 & 0.555 \\
    DB-MFIF (2024) & 0.8699 & 0.8247 & 0.6958 & 0.8663 & 0.8008 & 0.6647 & 1.0573 & 0.8367 & 0.7478 & 0.9637 & 0.8947 & 0.7770 & 2.700 \\
    FusionDiff (2024) & 0.8392 & 0.8225 & 0.6764 & 0.8557 & 0.8084 & 0.6132 & 0.9572 & 0.8310 & 0.7349 & 0.9526 & 0.8913 & 0.7220 & 26.915 \\
    MFFT (2024) & 1.1310 & 0.8393 & 0.6920 & 0.9336 & 0.8422 & 0.6865 & 1.1533 & 0.8433 & 0.7294 & 0.9523 & 0.8714 & 0.7511 & 1.769 \\
    SD-Fuse (2025) & 0.9378 & 0.8282 & 0.7126 & 0.9274 & 0.8553 & 0.6978 & 1.1233 & 0.8411 & 0.7601 & 0.9830 & 0.9100 & 0.7945 & 7.712 \\
    DMANet (2025) & 1.1513 & 0.8400 & 0.7358 & 0.9506 & 0.8659 & 0.7276 & 1.1876 & 0.8454 & 0.7591 & 0.9853 & 0.9083 & 0.8054 & 4.065 \\
    MCCSR (2025) & \underline{1.1768} & \underline{0.8418} & 0.7358 & \underline{0.9789} & \underline{0.8846} & \underline{0.7497} & \underline{1.1920} & \underline{0.8459} & 0.7619 & \textbf{0.9890} & 0.9106 & 0.8084 & 0.039 \\
    \textbf{CSNet} & \textbf{1.1858} & \textbf{0.8434} & \textbf{0.7366} & \textbf{0.9802} & \textbf{0.8888} & \textbf{0.7561} & \textbf{1.1924} & \textbf{0.8469} & 0.7592 & \underline{0.9882} & 0.9094 & \textbf{0.8094} & 1.893 \\
    \bottomrule
  \end{tabular}
  }
  \label{table 1}
\end{table*}
%-------------------------------------------------------------------------

%-------------------------------------------------------------------------

\subsection{Similarity Selection Strategy}
Moreover, boundary regions are typically narrow and occupy only a small fraction of the entire image. 
Although the reconstructed image $\boldsymbol{I}_{C}$ is beneficial for restoring ambiguous focused-defocused transitions, directly adopting its content over all boundary regions may disrupt the structural and gradient consistency with the source images, thereby introducing unnatural artifacts.
In addition, the focus maps estimated by CCAM, as well as the derived boundary map, are not always sufficiently fine-grained. 
Despite their reliability in large-scale region separation, irregular predictions such as jagged artifacts and local discontinuities may still appear around challenging edges around different regions.
This suggests that some pixels better suited for direct preservation from the source images may be incorrectly assigned to the boundary regions. 
To address these issues, we propose a Similarity Selection Strategy to further refine boundary pixels by adaptively selecting among the source images and the reconstructed image.

Notably, the initial clear image $\boldsymbol{I}_{C}$ integrates complementary information from the source images via cross-reconstruction. To approximate the AIF output, the reconstruction prioritized focused features while discarding blurry ones. 
At a given location, if $\boldsymbol{I}_{C}$ shows a strong bias toward one of the $\boldsymbol{I}_{A}$ or $\boldsymbol{I}_{B}$, as indicated by significantly higher feature similarity, it implies that this source image contributes the majority of focused information during reconstruction. 
The more effective features $\boldsymbol{I}_{C}$ inherits from a particular source image, the closer their representations become.
When this tendency exceeds a predefined threshold, directly selecting the pixel from the corresponding source image will better preserves fidelity and consistency in the fused output, while reducing the impact of Defocus Spread Effect (DSE).

Inspired by perceptual metrics aligned with human vision (LPIPS), we adopt a perceptual distance to characterize similarity. Multi-scale features are extracted via a pretrained VGG-16 network, and similarity is quantified as the inverse Euclidean distance between the resulting feature embeddings:
\begin{equation}
\begin{aligned}
\boldsymbol{S}_{A} &= 1 / (\sum_{l}\dfrac{1}{H_lW_l} \sum_{h,w} \left\| \left( \hat{\phi}^l(\boldsymbol{I_A}) - \hat{\phi}^l(\boldsymbol{I_C}) \right) \right\|_2^2 \\
\boldsymbol{S}_{B} &= 1 / (\sum_{l}\dfrac{1}{H_lW_l} \sum_{h,w} \left\| \left( \hat{\phi}^l(\boldsymbol{I_B}) - \hat{\phi}^l(\boldsymbol{I_C}) \right) \right\|_2^2
\end{aligned},
\end{equation}
where $\hat{\phi}^l(\cdot)$ denotes the unit-normalized feature map extracted from layer $l$ of the network, $(H_l, W_l)$ represents the spatial resolution of the feature map at layer $l$. Through comparing the calculation results, we obtain the similarity image:
\begin{equation}
\boldsymbol{I}_S(x,y)\!=\!\begin{cases} 
    \boldsymbol{I}_{A}(x,y), & \boldsymbol{S}_{A}(x,y) > \theta \cdot \boldsymbol{S}_{B}(x,y)\\
    \boldsymbol{I}_{B}(x,y), & \boldsymbol{S}_{B}(x,y) > \theta \cdot \boldsymbol{S}_{A}(x,y)\\
    \boldsymbol{I}_{C}(x,y), & else
\end{cases},
\end{equation}
where $\theta$ is a threshold to balance the fidelity and visual quality, setting to $1.2$ in our method.

Finally, we perform pixel-wise multiplication between $\boldsymbol{I}_{A}$, $\boldsymbol{I}_{B}$, $\boldsymbol{I}_{S}$ and corresponding masks to fuse the AIF output $\boldsymbol{I}_{F}$:
\begin{equation}
\boldsymbol{I}_F = \boldsymbol{I}_A \odot\boldsymbol{M}_A + \boldsymbol{I}_B \odot \boldsymbol{M}_B + \boldsymbol{I}_S \odot \boldsymbol{M}_C .
\end{equation}

\subsection{Loss Function}
The training loss function of our network consists of two components: the focus map loss $\mathcal{L}_M$ and the reconstruction loss $\mathcal{L}_R$. We use a synthetic dataset for training, providing the ground truths of focus maps $\boldsymbol{M}_{AG}$, $\boldsymbol{M}_{BG}$ and AIF result $\boldsymbol{I}_{G}$. 

The focus map loss is defined as:
\begin{equation}
\mathcal{L}_M = BCE(\boldsymbol{M}_{A}, \boldsymbol{M}_{AG}) + BCE(\boldsymbol{M}_{B}, \boldsymbol{M}_{BG}),
\end{equation}
where $BCE(\cdot,\cdot)$ denotes the Binary Cross Entropy loss. This term supervises the network to accurately localize focused regions in each source image, ensuring that the predicted focus maps faithfully reflect the true focus distribution and provide reliable guidance for boundary refinement.

The reconstruction loss is formulated as:
\begin{equation}
\mathcal{L}_R = \|\boldsymbol{I}_C - \boldsymbol{I}_G\|_{1} + \mu \cdot (1 - SSIM(\boldsymbol{I}_C, \boldsymbol{I}_G)),
\end{equation}
where $|\cdot|_1$ represents the L1 loss and $1 - SSIM(\cdot,\cdot)$ denotes the Structural Similarity Index Measure (SSIM) loss, they enforces fidelity of the reconstructed image $\boldsymbol{I}_C$ to the ground truth AIF image, promoting both pixel-level accuracy and perceptual consistency.

The total loss combines the two components as:
\begin{equation}
\mathcal{L} = \lambda_1 \cdot \mathcal{L}_M + \lambda_2 \cdot \mathcal{L}_R,
\end{equation}
$\mu, \lambda_1, \lambda_2$ are weights to balance the contributions of different terms, which are set as $[0.5, 1.0, 0.5]$ in our training.

%-------------------------------------------------------------------------
\begin{table*}[ht]
  \centering
  \caption{Quantitative comparison on the MFI-WHU and SIMIF datasets.}
  \normalsize
  \setlength{\heavyrulewidth}{1.2pt}
  \setlength{\lightrulewidth}{1pt}  
  \renewcommand{\arraystretch}{1.2}
  \resizebox{\textwidth}{!}{
  \begin{tabular}{l|cccccc|cccccc|c}
    \toprule
    \multirow{2}{*}{\textbf{Method}} & \multicolumn{6}{c|}{\textbf{MFI-WHU}} & \multicolumn{6}{c|}{\textbf{SIMIF}} & \multirow{2}{*}{\textbf{Time (s)}} \\
    \cmidrule(lr){2-7} \cmidrule(lr){8-13}
     & $Q_{MI}$ & $Q_{NCIE}$ & $Q_{G}$ & $Q_{Y}$ & $Q_{C}$ & $Q_{CB}$ & $Q_{MI}$ & $Q_{NCIE}$ & $Q_{G}$ & $Q_{Y}$ & $Q_{C}$ & $Q_{CB}$ & \\
    \midrule
    CSR & 0.9637 & 0.8307 & 0.7049 & 0.9151 & 0.8673 & 0.7558 & 1.0954 & 0.8420 & 0.7335 & 0.8749 & 0.8526 & 0.7310 & 94.732 \\
    DSIFT & 1.2209 & 0.8497 & 0.7321 & 0.9887 & \underline{0.9113} & \underline{0.8270} & 1.3120 & \underline{0.8562} & 0.7764 & 0.9840 & 0.9192 & 0.8347 & 1.166 \\
    MFM & 1.1791 & 0.8462 & 0.7324 & 0.9882 & 0.9106 & 0.8260 & 1.3002 & 0.8554 & \textbf{0.7783} & \underline{0.9849} & \textbf{0.9213} & 0.8327 & 0.104 \\
    SVDDCT & 1.2240 & 0.8501 & 0.7285 & \underline{0.9892} & 0.9808 & 0.8240 & 1.3120 & 0.8560 & 0.7762 & 0.9786 & 0.9152 & 0.8185 & 0.098 \\
    IFCNN & 0.8993 & 0.8275 & 0.6929 & 0.9404 & 0.8802 & 0.7367 & 1.0505 & 0.8403 & 0.7387 & 0.9211 & 0.8787 & 0.7364 & 0.084 \\
    SESF & 1.1878 & 0.8471 & 0.7294 & 0.9855 & 0.9088 & 0.8166 & 1.2995 & 0.8553 & 0.7773 & 0.9807 & 0.9159 & 0.8306 & 0.285 \\
    GACN & 1.2084 & 0.8489 & 0.7287 & 0.9889 & 0.9110 & 0.8241 & 1.3068 & 0.8557 & 0.7772 & 0.9845 & 0.9196 & \underline{0.8328} & 0.083 \\
    GRFusion & 1.2134 & 0.8486 & 0.7269 & 0.9848 & 0.9102 & 0.8212 & 1.2898 & 0.8543 & 0.7765 & 0.9771 & 0.9166 & 0.8267 & 0.619 \\
    MUFusion & 0.7449 & 0.8207 & 0.6107 & 0.8608 & 0.8155 & 0.6480 & 0.8876 & 0.8316 & 0.6674 & 0.8387 & 0.8139 & 0.6321 & 0.076 \\
    DB-MFIF & 1.0693 & 0.8380 & 0.7196 & 0.9466 & 0.8829 & 0.7818 & 1.1510 & 0.8463 & 0.7484 & 0.9157 & 0.8672 & 0.7612 & 0.061 \\
    FusionDiff & 0.9923 & 0.8315 & 0.7112 & 0.9401 & 0.8789 & 0.7937 & 1.1339 & 0.8439 & 0.7362 & 0.8945 & 0.8698 & 0.7458 & 85.279 \\
    MFFT & 1.1974 & 0.8486 & 0.7053 & 0.9614 & 0.8815 & 0.7648 & 1.2751 & 0.8537 & 0.7529 & 0.9492 & 0.8787 & 0.7724 & 0.168 \\
    SD-Fuse & 1.1574 & 0.8445 & 0.7278 & 0.9862 & 0.9113 & 0.8179 & 1.2725 & 0.8544 & 0.7699 & 0.9801 & 0.9161 & 0.8292 & 0.082 \\
    DMANet & 1.2220 & 0.8497 & \underline{0.7309} & 0.9878 & 0.9093 & 0.8222 & 1.3130 & 0.8561 & 0.7758 & 0.9780 & 0.9127 & 0.8270 & 0.114 \\
    MCCSR & \underline{1.2246} & \underline{0.8502} & 0.7243 & 0.9891 & 0.9088 & 0.8227 & \underline{1.3151} & 0.8561 & 0.7754 & 0.9835 & 0.9175 & 0.8298 & 0.283 \\
    \textbf{CSNet} & \textbf{1.2256} & \textbf{0.8503} & \textbf{0.7328} & \textbf{0.9893} & \textbf{0.9114} & \textbf{0.8281} & \textbf{1.3166} & \textbf{0.8563} & \underline{0.7774} & \textbf{0.9859} & \underline{0.9201} & \textbf{0.8348} & 0.062 \\
    \bottomrule
  \end{tabular}
  }
  \label{table 2}
\end{table*}
%-------------------------------------------------------------------------

%-------------------------------------------------------------------------
\begin{figure*}[ht]  
  \centering  
  \includegraphics[width=1\textwidth]{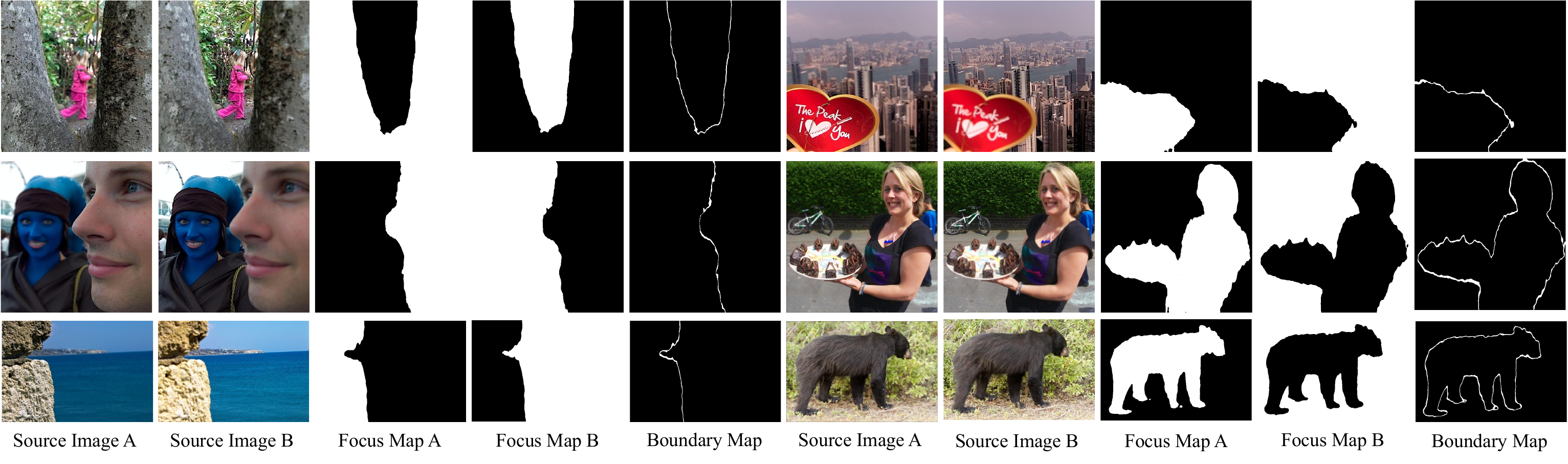}  
  \caption{Visualization of the focus maps and boundary maps generated by our CSNet. } 
  \vspace{-10pt}
  \label{fig:6}  
\end{figure*}
%-------------------------------------------------------------------------

%-------------------------------------------------------------------------
\begin{figure*}[t]  
  \centering  
  \vspace{-5pt}
  \includegraphics[width=1\textwidth]{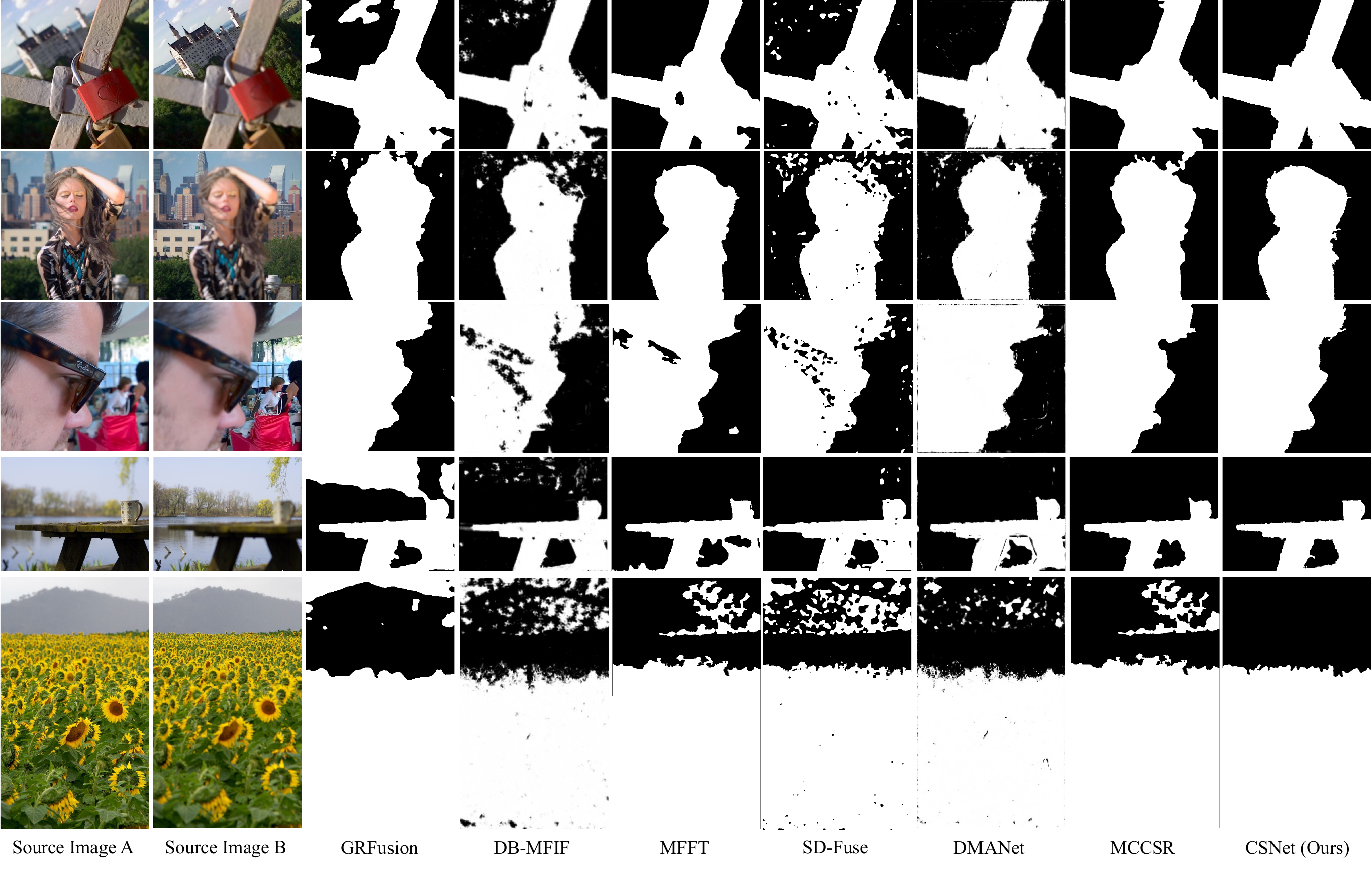}  
  \caption{Visualization of the focus maps on the Lytro and MFFW datasets.
  The white regions indicate the focused pixels in the source image $\boldsymbol{I}_{A}$.} 
  \label{fig:7}  
  \vspace{-5pt}
\end{figure*}
%-------------------------------------------------------------------------

%-------------------------------------------------------------------------
\subsection{Training datasets Generation}
One major challenge in MFIF lies in the lack of multi-focus datasets with real AIF ground truth (GT).
As a result, previous supervised methods are typically trained on synthetic datasets. 
The synthesis process generally involves two steps: dividing the focused regions of different source images and simulating defocus blur.
A part of previous works applied hand-labeled masks~\cite{li2020drpl} or random masks~\cite{wang2023multi} for division, but these often fail to capture the spatial distribution of regions in real photographic scenes. 
Some others select foreground images from matting datasets and composite them with different backgrounds, using $\alpha$ matting strategy to define precise boundaries~\cite{ma2020alpha}. 
However, such methods often result in large discrepancies between regions, leading to unnatural transitions. Typically, current methods utilize Gaussian kernels to simulate the effect of defocus blur.

Our method commits to identify the exactly focused regions in the source images and accurately locating the boundaries. Therefore, in the generation of training dataset, we not only segregate the physically plausible focused regions and simulate defocus blur, but also create smooth and natural boundaries to emulate realistic Defocus Spread Effect (DSE). 

Specifically, we select 15,000 image pairs from the COME dataset~\cite{zhang2021rgb}, each comprising an AIF RGB image $\boldsymbol{I}_F\in\mathbb{R}^{H \times W \times C}$ and a binary segmentation mask $\boldsymbol{m}\in[0, 1]^{H \times W}$, all of which are resized to a uniform resolution of $640 \times 360$. 
For every sample, we first divide the focused regions based on the segmentation mask. In real photographic scenes, the focused-defocused boundaries do not align with the actually physical boundaries of the objects due to the effects of DSE, instead of extending a certain distance outward from the physical boundaries. Therefore, we apply Gaussian blurring to the segmentation mask to create two focused region mask $\boldsymbol{m}_A$ and $\boldsymbol{m}_B\in[0, 1]^{H \times W}$:
\begin{gather}
\boldsymbol{m}_{A} = \boldsymbol{m} \otimes G(\sigma_{A}), \\
\boldsymbol{m}_{B} = (1-\boldsymbol{m}) \otimes G(\sigma_{B}),
\end{gather}
where $\boldsymbol{G}$ denotes the Gaussian kernel, and $\sigma_A$ and $\sigma_B$ are the standard deviation of the Gaussian kernels, which is randomly selected from the range $[1, 9]$, respectively. The kernel size is determined based on the standard deviation.

Subsequently, we employ Gaussian blurring to emulate defocus blur, and then combine the blurred image with the all-in-focus image according to masks to obtain the multi-focus image pair $\boldsymbol{I}_A, \boldsymbol{I}_B\in\mathbb{R}^{H \times W \times C}$:
\begin{gather}
\boldsymbol{I}_{A} = \boldsymbol{I}_{F} \odot \boldsymbol{m}_{A} + (\boldsymbol{I}_{F} \otimes G(\sigma_{A})) \odot (1- \boldsymbol{m}_{A}), \\
\boldsymbol{I}_{B} = \boldsymbol{I}_{F} \odot \boldsymbol{m}_{B} + (\boldsymbol{I}_{F} \otimes G(\sigma_{B})) \odot (1- \boldsymbol{m}_{B}).
\end{gather}
Finally, there are some remaining regions that are not covered in both $\boldsymbol{m}_A$ and $\boldsymbol{m}_B$, meaning they are not part of the focused regions in either $\boldsymbol{I}_A$ or $\boldsymbol{I}_B$. We label these areas as focused-defocused boundaries (FDB):
\begin{equation}
\boldsymbol{m}_{FDB} = \boldsymbol{1} - \boldsymbol{m}_{A} - \boldsymbol{m}_{B}.  
\end{equation}

\section{Experiments}
\subsection{Experimental Settings}

%-------------------------------------------------------------------------
\begin{figure*}[ht]  
  \centering  
  \includegraphics[width=0.99\textwidth]{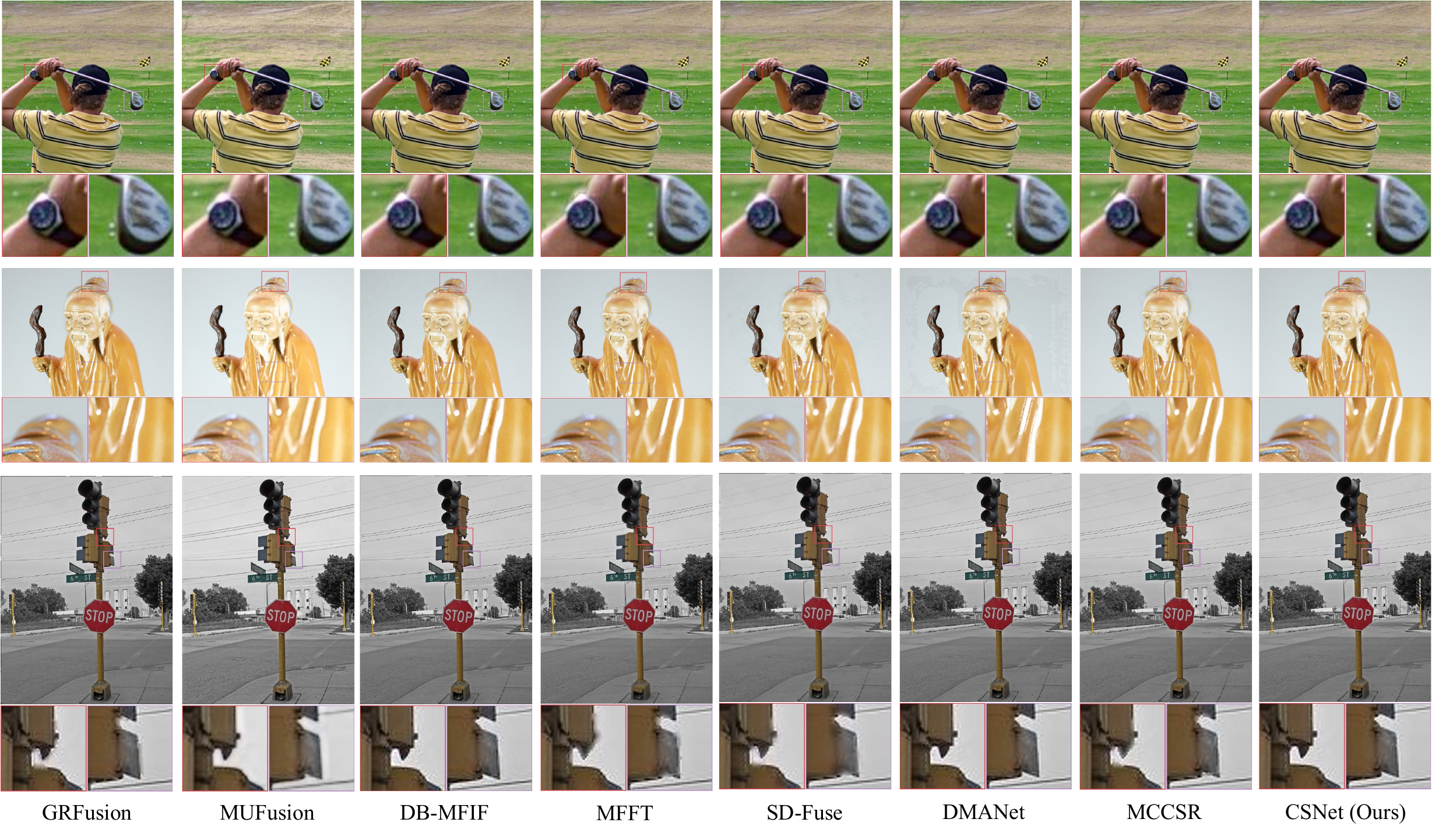}  
  \vspace{-10pt}
  \caption{Visualization of the fusion images from the Lytro, MFFW and MFI-WHU datasets. Two enlarged views are shown to reveal critical details.} 
  \vspace{-10pt}
  \label{fig:8}  
\end{figure*}
%-------------------------------------------------------------------------

%-------------------------------------------------------------------------
\begin{table*}[ht]
  \centering
  \caption{Ablation studies on the effect of the designs and components.}
  \vspace{-5pt}
  \normalsize
  \setlength{\heavyrulewidth}{1.2pt}
  \setlength{\lightrulewidth}{1pt}  
  \renewcommand{\arraystretch}{1.1}
  \resizebox{\textwidth}{!}{
  \begin{tabular}{l|cccccc|cccccc}
    \toprule
    \multirow{2}{*}{\textbf{Method}} & \multicolumn{6}{c|}{\textbf{MFFW}} & \multicolumn{6}{c}{\textbf{Lytro}} \\
    \cmidrule(lr){2-7} \cmidrule(lr){8-13}
     & $Q_{MI}$ & $Q_{NCIE}$ & $Q_{G}$ & $Q_{Y}$ & $Q_{C}$ & $Q_{CB}$ & $Q_{MI}$ & $Q_{NCIE}$ & $Q_{G}$ & $Q_{Y}$ & $Q_{C}$ & $Q_{CB}$ \\
    \midrule
    w/o CC & 1.1714 & 0.8418 & 0.7298 & 0.9701 & 0.8782 & 0.7479 & 1.1831 & 0.8447 & 0.7543 & 0.9806 & 0.9018 & 0.8027 \\
    w/o m-l & 1.1792 & 0.8430 & 0.7336 & 0.9762 & 0.8823 & 0.7492 & 1.1859 & 0.8452 & 0.7569 & 0.9843 & 0.9049 & 0.8058 \\
    w/o c.a. & 1.1786 & 0.8424 & 0.7340 & 0.9755 & 0.8789 & 0.7481 & 1.1868 & 0.8458 & 0.7564 & 0.9855 & 0.9069 & 0.8049 \\
    w/o alpha & 1.1833 & \underline{0.8437} & 0.7359 & \textbf{0.9810} & \underline{0.8882} & 0.7544 & \underline{1.1907} & 0.8468 & 0.7584 & \textbf{0.9883} & \underline{0.9087} & 0.8086 \\
    w/o s.a. & 1.1809 & \textbf{0.8438} & 0.7348 & 0.9770 & 0.8854 & 0.7518 & 1.1878 & 0.8461 & 0.7561 & 0.9859 & 0.9062 & 0.8057 \\
    \cmidrule(lr){1-13}
    ${I}_A$\&${I}_B$ & 1.1783 & 0.8426 & 0.7348 & 0.9762 & 0.8859 & 0.7502 & 1.1895 & 0.8463 & \textbf{0.7605} & 0.9860 & 0.9075 & 0.8054 \\
    ${I}_C$ & 1.1816 & 0.8431 & 0.7305 & 0.9735 & 0.8827 & \textbf{0.7568} & 1.1862 & 0.8453 & 0.7551 & 0.9848 & 0.9052 & \underline{0.8087} \\
    L1 & 1.1817 & 0.8430 & \textbf{0.7382} & 0.9796 & 0.8869 & 0.7542 & 1.1881 & 0.8461 & \underline{0.7598} & 0.9866 & 0.9068 & 0.8067 \\
    SSIM & 1.1841 & 0.8432 & \underline{0.7374} & 0.9747 & 0.8847 & 0.7539 & 1.1905 & \textbf{0.8471} & 0.7586 & 0.9868 & 0.9061 & 0.8072 \\
    cosine & \underline{1.1851} & 0.8436 & 0.7313 & 0.9763 & 0.8855 & 0.7551 & 1.1902 & 0.8467 & 0.7560 & 0.9857 & 0.9074 & 0.8079 \\
    \cmidrule(lr){1-13}
    w/o CC + ${I}_A$\&${I}_B$ & 1.1678 & 0.8408 & 0.7274 & 0.9678 & 0.8763 & 0.7471 & 1.1786 & 0.8438 & 0.7534 & 0.9785 & 0.8989 & 0.8011 \\
    w/o CC + ${I}_C$ & 1.1692 & 0.8419 & 0.7253 & 0.9665 & 0.8749 & 0.7503 & 1.1811 & 0.8449 & 0.7517 & 0.9812 & 0.9023 & 0.8024 \\
    \cmidrule(lr){1-13}
    \textbf{CSNet} & \textbf{1.1858} & 0.8434 & 0.7366 & \underline{0.9802} & \textbf{0.8888} & \underline{0.7561} & \textbf{1.1924} & \underline{0.8469} & 0.7592 & \underline{0.9882} & \textbf{0.9094} & \textbf{0.8094} \\
    \bottomrule
  \end{tabular}
  }
  \label{table 3}
  \vspace{-5pt}
\end{table*}
%-------------------------------------------------------------------------

\noindent \textbf{Testing datasets}
We evaluate the performance of MFIF on four benchmark datasets: Lytro~\cite{nejati2015multi}, MFFW~\cite{xu2020mffw}, MFI-WHU~\cite{zhang2021mff} and SIMIF~\cite{tsai2014standardimages}. 
The Lytro dataset contains 20 pairs of multi-focus images captured by a light field camera. 
The MFFW dataset includes 13 real image pairs with strong DSE. 
The MFI-WHU dataset is constructed using Gaussian blur and handmade decision map and consists of a larger scale with 120 pairs. 
The SIMIF dataset is composed of 12 pairs of high-resolution real images. The chosen datasets have different characteristics, thus can be utilized to examine the image fusion performance comprehensively.
\vspace{5pt}

%-------------------------------------------------------------------------
\noindent \textbf{Implementation details}
Our CSNet is implemented in PyTorch and trained on NVIDIA GeForce RTX 4090 GPUs, using 15,000 synthesized image pairs generated by our approach. We employ the AdamW optimizer, with a batch size of 8, a learning rate of $1\times10^{-4}$, and train for 50 epochs. 
During the inference phase, a small region removal strategy, common in the decision-based methods~\cite{zheng2025unfolding}, is used to refine decision maps. The threshold is set to $0.01 \times H \times W$, where $H \times W$ is the size of source images.
\vspace{5pt}

%-------------------------------------------------------------------------
\noindent \textbf{Competitive methods}
The performance of CSNet is evaluated against 15 representative MFIF approaches, $4$ of them are traditional methods: CSR~\cite{liu2016image}, DSIFT~\cite{liu2015multi}, MFM~\cite{ma2017multi} and SVDDCT~\cite{amin2017multi}; and others are based on deep learning: IFCNN~\cite{zhang2020ifcnn}, SESF~\cite{ma2021sesf}, GACN~\cite{ma2022end}, GRFusion~\cite{li2023generation}, MUFusion~\cite{cheng2023mufusion}, DB-MFIF~\cite{zhang2024exploit}, FusionDiff~\cite{li2024fusiondiff}, MFFT~\cite{zhai2024multi}, SD-Fuse~\cite{wang2025sd}, DMANet~\cite{quan2025multi} and MCCSR~\cite{zheng2025unfolding}. For all methods, we adopt their official open-source implementations and pre-trained weights for inference. 
This benchmark encompasses the mainstream MFIF methods developed over the past five years, providing an exhaustive and fair comparison.
\vspace{5pt}

%--------------------------------------------------------------
\noindent \textbf{Metrics}
We employ six popular objective metrics for the performance evaluation, including the information theory-based metrics $Q_{MI}$~\cite{hossny2008comments} and $Q_{NCIE}$~\cite{wang2005nonlinear}, the edge based similarity measurement $Q_{G}$~\cite{xydeas2000objective}, the image structural similarity-based metrics $Q_{Y}$~\cite{yang2008novel} and $Q_{C}$~\cite{cvejic2005similarity}, and the human
perception-inspired metrics $Q_{CB}$~\cite{chen2009new}. 
These metrics provide a comprehensive assessment of MFIF methods from multiple perspectives~\cite{zhang2021deep, liu2024rethinking}.

%-------------------------------------------------------------------------
\begin{figure*}[ht]  
  \centering  
  \vspace{-5pt}
  \includegraphics[width=1.0\textwidth]{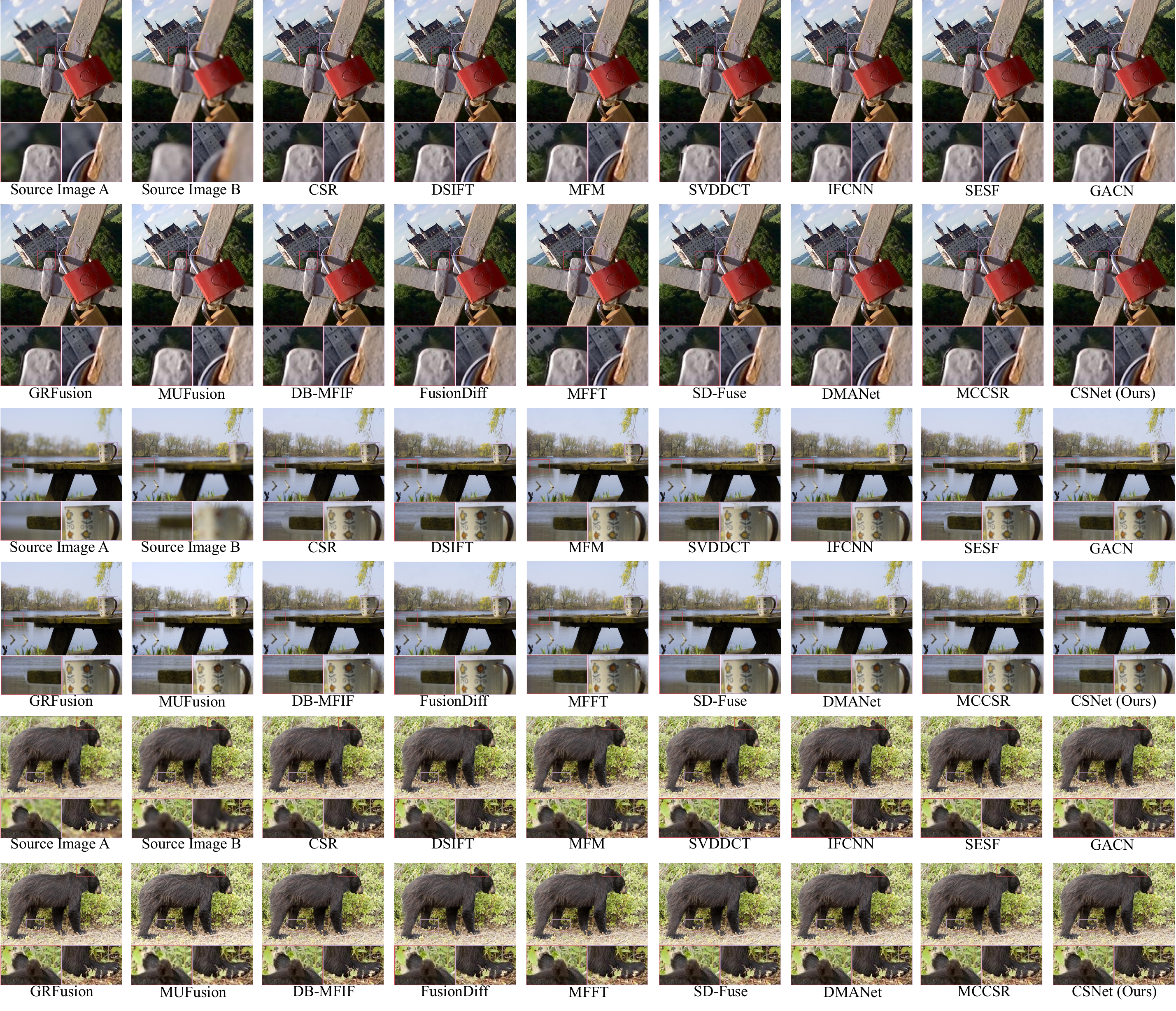}  
  \caption{Visualization of the fusion images from the Lytro, MFFW and MFI-WHU datasets.} 
  \vspace{-5pt}
  \label{fig:9}  
\end{figure*}
%-------------------------------------------------------------------------

\subsection{Performance Evaluation}
\noindent \textbf{Quantitative comparison}
The quantitative results on the classic MFIF benchmarks MFFW and Lytro are reported in Table~\ref{table 1}. 
On the MFFW dataset who introduces a strong Defocus Spreading Effect (DSE), CSNet exhibits obviously superior results on all evaluation metrics, highlighting the effectiveness of our approach in handling the boundaries between focused and defocused regions. 
Concurrently, our method demonstrates competitive performance on the Lytro dataset. However, since there is no consideration on the DSE in the Lytro dataset, our refinement strategy cannot be fully utilized, and CSNet does not achieve a significant advantage over previous methods. 

Table~\ref{table 2} presents the comparison on the MFI-WHU and SIMIF datasets. In the complex scenes with larger-scale of the WHU-MFI dataset and the high-resolution images of the SIMIF dataset, CSNet realizes the highest scores on almost all metrics with an absolute advantage. Evaluations on metrics demonstrates\ the strong generalization capability of our method across diverse types of datasets.

Besides, we observe that other methods struggle to consistently deliver well behavior across different datasets and metrics, which further underscores the challenges posed by the diverse characteristics of datasets and multidimensional evaluation. 
In summary, across the four listed datasets and six metrics for each, our method achieves a total 19 best and 3 second-place results, showing strong generalization and comprehensive performance. 
This is especially evident in the information theory-based and human visual system-based metrics, indicating that our fused images thoroughly preserve the salient information from the source images as well generate visually pleasing results.
\vspace{5pt}
%-------------------------------------------------------------------------

\noindent \textbf{Qualitative comparison}
To demonstrate the effectiveness of the proposed clarity contrast mechanism in accurately localizing the FDB, Figure~\ref{fig:6} presents several representative focus maps and boundary maps generated by our CSNet.

To more intuitively verify the capability of our method, Figure~\ref{fig:7} compares the focus maps for the source image $\boldsymbol{I}_{A}$ generated by CSNet against several state-of-the-art decision-based approaches. 
Note that the white regions in the focus maps indicate the focused pixels in $\boldsymbol{I}_{A}$.
Our focus maps more clearly distinguish focused regions with fewer voids, while maintaining structurally coherent transitions around the boundaries, without disrupting the physical contours of objects. 
In contrast, some methods exhibit obvious holes (GRFusion, MCCSR) in their focus maps, display considerable noise (SD-Fuse, DMANet) or produce coarse even fragmented boundaries (DB-MFIF), suggesting incorrect assessments of the focusing attributes.

As shown in Figure~\ref{fig:8}, we illustrate a visualization of the fusion results. The compared approaches encounter various issues.
For instance, GRFusion, MFFT, MCCSR suffer from significant edge ghosting artifacts, MUFusion introduces obvious color deviation, DB-MFIF, SD-Fuse, DMANet include textural noise and blurred details.
In contrast, our method avoids unnatural artifacts or chromatic distortion and accurately collect the focused regions as well as generating clear boundaries.
Specifically, for the first `golf' sample, we produce more accurate contours of watch and wrist as well sharper club boundaries. 
In the second `statue' case, CSNet avoids producing artifacts around the head while faithfully retaining the reflections on the torso. 
In the "traffic light" example, our method not only precisely extracts the focused regions on panel, but also restores sharp tips.
Through visualization and comparison, our approach identifies the focused regions with superior boundaries and displays pleasing visual quality. More extensive qualitative results are provided in Figure~\ref{fig:9}.
%-------------------------------------------------------------------------

\noindent \textbf{Computational complexity comparison}
The number of learnable model parameters and the inference time for one pair of input are also displayed in Table~\ref{table 1} and Table~\ref{table 2}. We conduct unified tests using random inputs of $512 \times 512$ resolution and measure the average end-to-end processing time from the source images to the fused result. For deep-learning based methods, inferences are performed on a single RTX 4090 GPU. For traditional methods, tests are implemented in the same MATLAB environment on CPU. 
Compared to the listed methods, CSNet maintains moderate computational complexity, suggesting that our performance gains arise primarily from elegant designs and strategies rather than complexity inflation.

%-------------------------------------------------------------------------
\begin{table*}[ht]
  \centering
  \caption{Sensitivity Analysis on the settings of hyperparameters.}
  \normalsize
  \setlength{\heavyrulewidth}{1.2pt}
  \setlength{\lightrulewidth}{1pt}  
  \renewcommand{\arraystretch}{1.1}
  \resizebox{\textwidth}{!}{
  \begin{tabular}{ccc|cccccc|cccccc}
    \toprule
    \multicolumn{3}{c|}{\textbf{Setting}} & \multicolumn{6}{c|}{\textbf{MFFW}} & \multicolumn{6}{c}{\textbf{Lytro}} \\
    \cmidrule(lr){1-15}
    \multicolumn{3}{c|}{$\theta$} & $Q_{MI}$ & $Q_{NCIE}$ & $Q_{G}$ & $Q_{Y}$ & $Q_{C}$ & $Q_{CB}$ & $Q_{MI}$ & $Q_{NCIE}$ & $Q_{G}$ & $Q_{Y}$ & $Q_{C}$ & $Q_{CB}$ \\
    \cmidrule(lr){1-3} \cmidrule(lr){4-9} \cmidrule(lr){10-15}
    \multicolumn{3}{c|}{$1.05$} & 1.1814 & 0.8429 & 0.7351 & 0.9798 & 0.8862 & 0.7533 & 1.1905 & 0.8468 & 0.7589 & 0.9876 & 0.9077 & \textbf{0.8096} \\
    \multicolumn{3}{c|}{$1.10$} & 1.1828 & 0.8430 & 0.7360 & \underline{0.9799} & 0.8879 & 0.7558 & 1.1911 & 0.8468 & 0.7588 & 0.9878 & 0.9083 & 0.8087 \\
    \multicolumn{3}{c|}{$1.15$} & 1.1842 & 0.8431 & \textbf{0.7368} & \underline{0.9799} & \underline{0.8884} & \underline{0.7562} & 1.1919 & \textbf{0.8470} & 0.7591 & 0.9879 & \textbf{0.9095} & 0.8092 \\
    \multicolumn{3}{c|}{$1.20$} & \textbf{1.1858} & \textbf{0.8434} & \underline{0.7366} & \textbf{0.9802} & \textbf{0.8888} & 0.7561 & \textbf{1.1924} & \underline{0.8469} & \textbf{0.7592} & \underline{0.9882} & \underline{0.9094} & \underline{0.8094} \\
    \multicolumn{3}{c|}{$1.25$} & \underline{1.1844} & \underline{0.8432} & 0.7355 & 0.9794 & 0.8876 & \textbf{0.7566} & \underline{1.1922} & \underline{0.8469} & 0.7590 & \textbf{0.9884} & \textbf{0.9095} & 0.8091 \\
    \multicolumn{3}{c|}{$1.30$} & 1.1835 & 0.8429 & 0.7342 & 0.9776 & 0.8878 & \textbf{0.7566} & 1.1908 & 0.8464 & 0.7574 & 0.9872 & 0.9088 & 0.8089 \\
    \cmidrule(lr){1-15}
    $\lambda_1$ & $\lambda_2$ & $\mu$ & $Q_{MI}$ & $Q_{NCIE}$ & $Q_{G}$ & $Q_{Y}$ & $Q_{C}$ & $Q_{CB}$ & $Q_{MI}$ & $Q_{NCIE}$ & $Q_{G}$ & $Q_{Y}$ & $Q_{C}$ & $Q_{CB}$ \\
    \cmidrule(lr){1-3} \cmidrule(lr){4-9} \cmidrule(lr){10-15}
    $1.0$ & $1.0$ & $0.5$ & 1.1821 & 0.8421 & 0.7342 & 0.9779 & 0.8871 & 0.7553 & 1.1891 & 0.8462 & 0.7573 & 0.9858 & 0.9072 & 0.8080 \\
    $1.0$ & $1.0$ & $1.0$ & 1.1826 & 0.8423 & \underline{0.7361} & 0.9788 & 0.8874 & 0.7559 & \underline{1.1914} & \underline{0.8467} & 0.7582 & \underline{0.9879} & 0.9078 & 0.8086 \\
    $0.5$ & $1.0$ & $0.5$ & \underline{1.1839} & \underline{0.8427} & 0.7358 & \textbf{0.9813} & \underline{0.8882} & \textbf{0.7566} & 1.1903 & 0.8463 & \textbf{0.7597} & 0.9878 & \underline{0.9084} & \underline{0.8089} \\
    $1.0$ & $0.5$ & $0.0$ & 1.1772 & 0.8416 & 0.7309 & 0.9762 & 0.8855 & 0.7521 & 1.1847 & 0.8454 & 0.7560 & 0.9841 & 0.9061 & 0.8067 \\
    $1.0$ & $1.0$ & $0.0$ & 1.1795 & 0.8424 & 0.7337 & 0.9785 & 0.8869 & 0.7542 & 1.1868 & 0.8456 & 0.7578 & 0.9846 & 0.9066 & 0.8072 \\
    $1.0$ & $0.5$ & $0.5$ & \textbf{1.1858} & \textbf{0.8434} & \textbf{0.7366} & \underline{0.9802} & \textbf{0.8888} & \underline{0.7561} & \textbf{1.1924} & \textbf{0.8469} & \underline{0.7592} & \textbf{0.9882} & \textbf{0.9094} & \textbf{0.8094} \\
    \bottomrule
  \end{tabular}
  }
  \label{table 4}
  \vspace{-15pt}
\end{table*}
%-------------------------------------------------------------------------

\subsection{Ablation Studies}
To quantitatively validate the effectiveness of the designs, strategies and hyperparameters settings in the CSNet, we conduct a series of ablation studies and report the results on the Lyrto and MFFW datasets.
\vspace{5pt}
%-------------------------------------------------------------------------

\noindent \textbf{Effect of the Clarity Contrast Attention Module (CCAM)}
We investigate the contributions of key components in CCAM through the ablations summarized in Table~\ref{table 3}. From the complete module, we progressively remove or modify elements:

\textbf{(1) w/o CC:} The clarity contrast is removed, and the source images are processed independently without mutual modulation.  
\textbf{(2) w/o m-l:} Clarity contrast is applied only at the deepest layer.  
Both modifications result in noticeable performance degradation across all metrics, demonstrating that establishing multi-level interactions between source images is critical for accurate focus discrimination.

\textbf{(3) w/o c.a.:} Channel attention is replaced with ordinary convolutional layers.  
\textbf{(4) w/o $\alpha$:} The residual connection within the channel attention is removed.  
\textbf{(5) w/o s.a.:} Spatial attention is discarded.  
The observed declines in quantitative scores indicate that each component plays an important role in enhancing clarity-aware feature representation and ensuring robust boundary localization.
\vspace{5pt}

%-------------------------------------------------------------------------
\noindent \textbf{Effect of the Similarity Selection Strategy (SSS)}
In CSNet, boundary pixels are adaptively selected from the appropriate source images by comparing similarity. To illustrate the rationale behind this design, we compare several alternative strategies in Table~\ref{table 3}.

\textbf{(1)} $\boldsymbol{I}_A$\&$\boldsymbol{I}_B$: Without similarity comparison and purposeful selection, we force the focus maps to cover all pixels of the image. This leads to degraded performance near boundaries, where both source images may contain blurred content. 
\textbf{(2)} $\boldsymbol{I}_C$: The boundary pixels are directly taken from the reconstructed image $\boldsymbol{I}_{C}$. 
Due to its finite representation capacity, $\boldsymbol{I}_{C}$ cannot perfectly reconstruct all focused pixels from the source images. Directly using its content can compromise fidelity and consistency, resulting in artifacts and lower performance.

\textbf{(3) L1, (4) SSIM, (5) cosine:} We further replace the proposed perceptual-distance-based criterion with common similarity measures. While these approaches provide a basic guidance for pixel assignment, their overall performance is inferior, indicating that the perceptual distance can better balance structural consistency and boundary refinement.
\vspace{5pt}

\noindent \textbf{Collaborative Effect of the CCAM and SSS}
The ablation results in Table~\ref{table 3} show that the CCAM and the SSS work synergistically to improve fusion performance. 
Removing CCAM while applying $\boldsymbol{I}_A$\&$\boldsymbol{I}_B$ or only $\boldsymbol{I}_C$ at the boundaries results in a significant performance drop across all metrics. 
Importantly, the performance drop in these combined ablations exceeds that of removing either module individually, demonstrating their complementarity. 
CCAM establishes multi-level focus interactions and delineates boundaries, while SSS adaptively assigns pixels within these regions. 
Disabling either impairs the other, leading to compounded errors in focus discrimination and boundary fidelity. 
This mutual reinforcement highlights that integrating multi-level clarity contrast with similarity-guided boundary selection is crucial for optimal fusion performance.

Overall, the ablation results confirm that both inter-image interactions and carefully designed selection strategy are essential for the high performance of CSNet. The consistent drop across Lytro and MFFW datasets further validates the generality and necessity of these design choices.

\subsection{Sensitivity Analysis on the Settings of Hyperparameters}
To assess the sensitivity and robustness of CSNet to hyperparameter settings, we conduct ablation studies on key parameters including the similarity threshold $\theta$  in the fusion strategy and the loss function weights $(\lambda_1, \lambda_2, \mu)$.
The quantitative results on two datasets are summarized in Table~\ref{table 4}.

For the threshold $\theta$, CSNet exhibits stable performance across a range of values from $1.05$ to $1.30$ on both datasets. The results vary slightly under different thresholds, while $\theta=1.20$ achieves the best performance on most metrics, suggesting that it offers an appropriate trade-off between reliable boundary selection and focused detail preservation.

For the loss weights $(\lambda_1, \lambda_2, \mu)$, the results show that both the focus map supervision and the reconstruction constraint are necessary for stable fusion performance. 
Removing the SSIM term by setting leads to noticeable degradation, while CSNet shows limited performance fluctuations under different weighting schemes.
The setting $(\lambda_1,\lambda_2,\mu)=(1.0,0.5,0.5)$ achieves the strongest performance, indicating that a moderate balance among the loss components.

%-------------------------------------------------------------------------

%-------------------------------------------------------------------------
\begin{figure}[ht]  
  \centering  
  \vspace{-5pt}
  \includegraphics[width=0.488\textwidth]{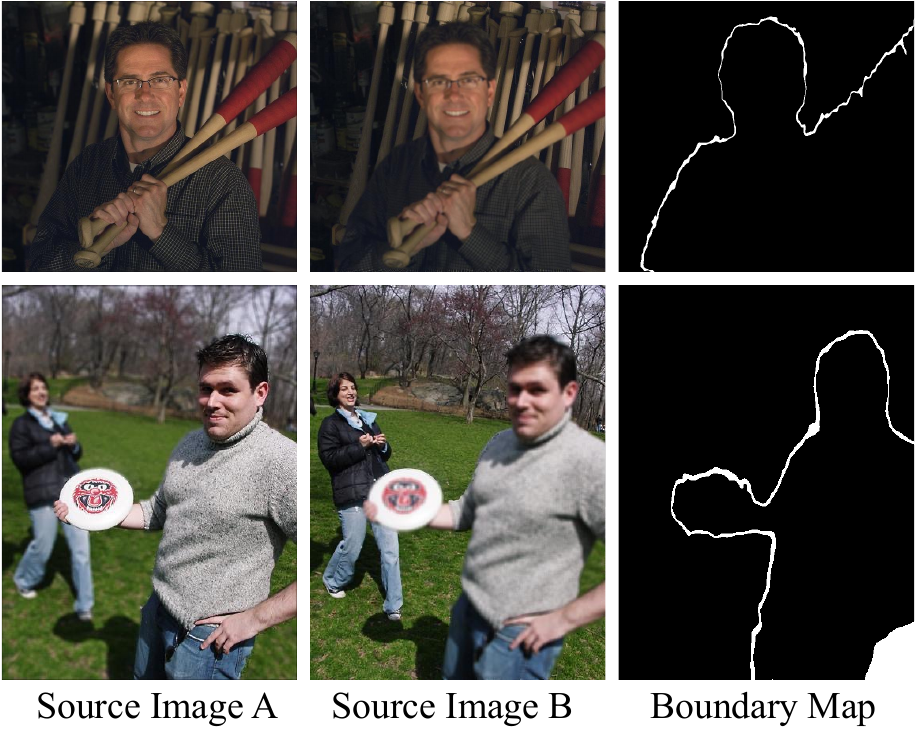}  
  \caption{Representative failure cases of the proposed method.} 
  \vspace{-5pt}
  \label{fig:10}  
\end{figure}
%-------------------------------------------------------------------------

\subsection{Failure Cases Analysis}
To further assess the capabilities and limitations of CSNet, we present several representative failure cases in Figure~\ref{fig:10}. These examples highlight scenarios in which the boundary maps produced by our model do not fully reach the expected precision from real and synthetic datasets.

In particular, under extreme conditions, such as when focused regions are manually annotated or the distinction between focused and defocused areas is subtle. The estimated focus maps may exhibit minor inaccuracies. As a result, the derived boundary maps, while correctly capturing most transition regions, may occasionally miss some boundaries or slightly extend into non-boundary areas. Although such errors do not lead to complete fusion failure, they can slightly compromise the overall quality of the fused result.

It is important to note that these failure cases are rare and mostly occur under challenging scenarios. However, they highlight meaningful directions for further improving the robustness of our approach. For example, introducing segmentation or depth prior into focus map learning and incorporating additional contextual cues could help better resolve ambiguous transitions and enhance boundary fidelity.

%-------------------------------------------------------------------------
\section{Conclusion}
In this paper, we propose a Clarity Contrast and Similarity Selection Network (CSNet), to establish an explicit interaction between the source images for MFIF.
Specifically, by contrasting the clarity differences, we mutually identify the focused regions in source images to generate two focus maps, along with locating the boundaries, which are further refined by our Similarity Selection Strategy.
Through this interactive approach, CSNet preserves focused regions and reconstructs coherent boundaries to fuse an all-in-focus output.
Experimental evaluations on four benchmark datasets validate the superiority of our method.

\bibliographystyle{IEEEtran}
\vfill
\bibliography{references}

\end{document}